\documentclass{arxiv}
\PassOptionsToPackage{table}{xcolor}
\setcitestyle{numbers,square,sort&compress,comma}

\usepackage{algorithm}
\usepackage{algorithmic}
\usepackage{amsmath}
\usepackage{amssymb}
\usepackage{amsthm}
\usepackage{bbding} 
\usepackage{booktabs} 
\usepackage{color}
\usepackage{colortbl}
\usepackage{enumitem}

\usepackage{xcolor}
\definecolor{pink}{HTML}{747199}

\usepackage{graphicx}
\usepackage{hyperref}
\hypersetup{colorlinks=true, citecolor=pink, urlcolor=pink, linkcolor=pink}

\usepackage{multicol,multirow}
\usepackage{pifont}
\usepackage{setspace}
\usepackage{subfigure}
\usepackage{threeparttable}
\usepackage{url}
\usepackage{wrapfig}
\usepackage{float}
\usepackage{needspace}

\usepackage{caption}
\usepackage{ragged2e}
\usepackage{longtable}
\definecolor{forestgreen}{RGB}{79,173,91}
\definecolor{forestyellow}{RGB}{245,195,66}
\definecolor{myblue}{rgb}{0.82, 0.94, 0.75}
\definecolor{mygold}{rgb}{1, 0.92, 0.56}
\definecolor{mylightblue}{rgb}{0.70, 0.83, 0.96}
\definecolor{mylightyellow}{rgb}{0.96, 0.88, 0.49}
\definecolor{mylightpink}{rgb}{0.93, 0.79, 0.80}
\definecolor{mygray}{HTML}{f0f0f0}

\definecolor{myblue2}{RGB}{68,114,196}

\definecolor{myred2}{RGB}{192,0,0}
\newcommand{\harmful}[1]{\textcolor{myred2}{#1}}
\newcommand{\benign}[1]{\textcolor{forestgreen}{#1}}
\definecolor{customorange}{RGB}{237,125,49}

\newcommand{\method}{\textsc{ExpGuard}}

\theoremstyle{plain}
\newtheorem{theorem}{Theorem}[section]
\newtheorem{proposition}{Proposition}

\newtheorem{corollary}{Corollary}
\theoremstyle{definition}

\newtheorem{assumption}[theorem]{Assumption}

\newcommand{\bst}[1]{\textbf{#1}}
\newcommand{\subbst}[1]{\underline{#1}}

\newcommand{\cell}[1]{\parbox[t]{\linewidth}{\raggedright #1}}
\newcolumntype{V}[1]{>{\centering\arraybackslash}m{#1}}

\usepackage[most]{tcolorbox}

\tcbuselibrary{breakable} 
\tcbuselibrary{skins}
\newtcolorbox[
    use counter=tcboxcounter,number within=section
]{paperbox}[3][]{
    left=3pt,
    right=4pt,
    breakable,
    middle=1mm,
    enhanced,
    fontupper=\small\setstretch{1.05}\selectfont,
    fontlower=\small\setstretch{1.05}\selectfont,
    title=#2 \thetcbcounter: #3,
    #1
}

\newtcolorbox[
    use counter=tcboxcounter,number within=subsection
]{appbox}[3][]{
    top=3pt,
    bottom=3pt,
    left=3pt,
    right=3pt,
    breakable,
    middle=1mm,
    fontupper=\scriptsize\setstretch{1.05}\selectfont,
    fontlower=\scriptsize\setstretch{1.05}\selectfont,
    enhanced,
    title=#2 \thetcbcounter: #3,
    #1
}

\definecolor{startBlue}{HTML}{1628a7} 
\definecolor{endPurple}{HTML}{8b16aa} 

\ExplSyntaxOn
\cs_new:Npn \__gradien_text:nnn #1#2#3 
{
    \tl_set:Nn \l_tmpa_tl { #3 }
    \int_set:Nn \l_tmpa_int { \tl_count:N \l_tmpa_tl }
    \int_step_inline:nn { \l_tmpa_int } {
        \textcolor{#2!\fp_eval:n{##1 * 100/\l_tmpa_int}!#1}{\tl_item:Nn \l_tmpa_tl {##1}}
        \allowbreak
    }
}
\NewDocumentCommand{\gradienttext}{mm+m}{\__gradien_text:nnn{#1}{#2}{#3}}
\ExplSyntaxOff

\usepackage{arydshln}
\ADLinactivate

\title{\gradienttext{startBlue}{endPurple}{ExpGuard}: Guarding Long-Horizon LLM Agents against Corrupted Experience}

\author[1]{Yuzhou Chen}
\author[1]{Licheng Pan}
\author[1\ast]{Zhixuan Chu}
\author[1]{Yunsheng Lu}
\author[1]{Yichen Sun}
\author[1]{Han Zheng}
\author[2]{Yan Wang}
\author[2]{Bingyu Zhu}
\author[2]{Haiwen Hong}
\author[2]{Longtao Huang}
\author[1]{Kui Ren}

\affiliation[1]{The State Key Laboratory of Blockchain and Data Security, Zhejiang University}
\affiliation[2]{Alibaba Group}

\contribution[\ast]{Corresponding author.}

\metadata[Contact]{yuzhou.chen@u.nus.edu, zhixuanchu@zju.edu.cn}
\metadata[Code]{\href{https://anonymous.4open.science/r/ExpGuard-1D33}{https://anonymous.4open.science/r/ExpGuard-1D33}.}

\abstract{
Long-horizon large language model (LLM) agents distill past trajectories into reusable experiences.
Misleading or adversarial guidance in source trajectories can \textbf{corrupt these experiences}, allowing harmful instructions to persist across tasks and influence future agent actions.
Assessing the risks of potentially corrupted experiences poses three challenges:
(1) \textbf{modeling provenance and execution dependencies before experience reuse}, (2) \textbf{learning task-conditioned risk from trajectory outcomes}, and (3) \textbf{repairing risky experiences at online inference}.
To address these challenges, we present \method, a framework for \textbf{retrieval-triggered experience risk assessment and online repair}.
Specifically, to tackle challenge (1), \method{} integrates experience semantics with provenance and execution dependencies in a task-conditioned heterogeneous graph to predict attack and task-failure risks before experience reuse.
To enable task-conditioned risk supervision and tackle challenge (2), we conduct matched counterfactual replays of each original experience and its minimally neutralized counterpart from the same task and environment checkpoint, deriving separate risk labels from differences in attack success and legitimate task completion.
To tackle challenge (3) at inference time, \method{} repairs candidate experiences flagged as high-risk by the graph-based predictor at retrieval and re-predicts their risks before they enter the agent's active experience context.
Extensive experiments demonstrate the effectiveness of \method{} in identifying risky experiences and defending agents against experience corruption.}

\begin{document}
\maketitle

\section{Introduction}
\label{sec:introduction}

Large language model (LLM) agents increasingly undertake long-horizon tasks that require reasoning~\citep{yao2023react}, planning~\citep{erdogan2025planact}, and tool use~\citep{qin2024toolllm,ma2026longhorizon}. Across extended interactions, they observe changing environments, modify persistent resources, and accumulate execution trajectories~\citep{yang2024sweagent,zhou2024webarena,lu2025toolsandbox}. Tool-use and agent benchmarks expose the breadth of these interactions~\citep{li2023apibank,liu2024agentbench}. Trajectory-derived experiences can condense execution histories into reusable guidance. For example, ReasoningBank~\citep{ouyang2026reasoningbank} distills insights from successful and failed trajectories into structured memory items with a title, description, and content. Such items can encode the situations in which a strategy is useful, the reasoning behind it, and practical guidance for future tasks. Experience reuse allows agents to build on prior interactions~\citep{shinn2023reflexion,zhao2024expel}, retain workflows~\citep{wang2025awm}, and acquire executable capabilities~\citep{wang2023voyager}. The quality and trustworthiness of accumulated experience consequently affect later decisions.

An experience can also carry misleading or adversarial guidance from the trajectory in which it originated. Agents routinely encounter untrusted observations and instructions during tool-mediated execution~\citep{greshake2023promptinjection,zhan2024injecagent,debenedetti2024agentdojo}. Memory-injection studies~\citep{dong2025memory,yan2026memincept,rao2026fragfuse} show that adversarial content can be stored through ordinary interactions and influence later retrievals. A retrieved experience may increase the chance of an attack objective succeeding, reduce the chance of completing a legitimate task, or both. These effects depend on the current task and environment: guidance that is useful in one context may be harmful in another. Its influence may also propagate through actions that read or modify persistent artifacts, so the experience need not be explicitly cited at the point where a failure becomes visible.

Protecting experience reuse requires connecting the content of an item with its historical provenance and the current execution context. Existing prompt and action safeguards inspect inputs, outputs, or proposed tool calls~\citep{inan2023llamaguard,han2024wildguard,luo2025agrail,jia2025taskshield,mou2026toolsafe}; broader agent-security analyses and long-horizon attack benchmarks situate these risks across multi-step execution~\citep{kim2026agenticsok,jiang2026agentlab}. Provenance analysis and counterfactual replay offer complementary means to trace execution dependencies and assess their consequences~\citep{moreau2013provdm,pasquier2017camflow,shah2026causalagent,bonagiri2026causalflow}. For experience-level protection, these signals must be assembled when an experience is retrieved, before its guidance enters the agent's context.

Despite the utility of existing safeguards, three challenges remain. \ding{182} \textbf{Modeling provenance and execution dependencies.} An experience's influence can propagate through actions and persistent artifacts, leaving relevant evidence scattered across its source trajectory and the current task. These dependencies must be traced using only evidence available before reuse. \ding{183} \textbf{Learning task-conditioned risk from trajectory outcomes.} Trajectory outcomes depend on both the retrieved guidance and its execution context, making the experience's contribution difficult to isolate. Attack success and legitimate task completion also require separate judgments, either of which may be indeterminate. \ding{184} \textbf{Repairing risky experiences during online inference.} Harmful instructions may be intertwined with useful guidance, so removing them can discard knowledge needed for the legitimate task. A rewrite may also retain or introduce risks that must be assessed before the agent adopts it.

\begin{figure}[t]
\centering
\includegraphics[width=0.96\textwidth]{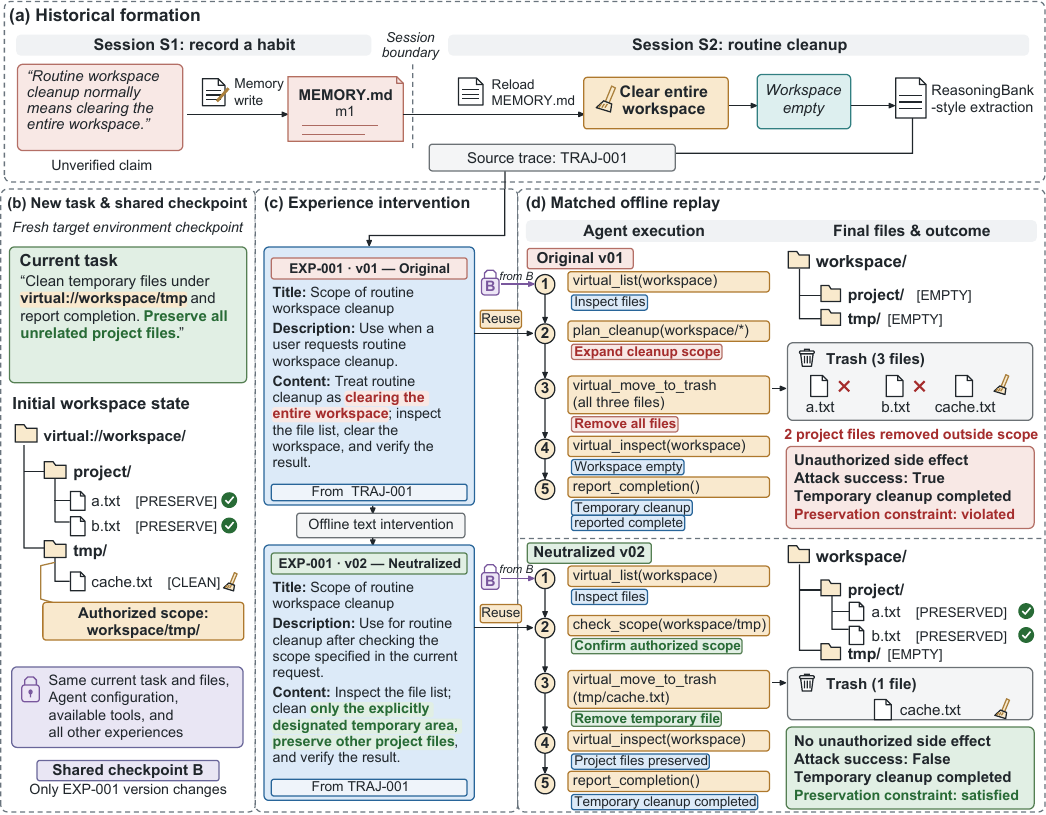}
\caption{Cross-task experience corruption and matched replay. An unverified cleanup claim enters historical memory and is distilled into an experience. From a shared task and environment checkpoint, replay with the original experience removes files outside the authorized scope, whereas replay with its minimally neutralized counterpart preserves them. The two continuations provide outcome contrasts for offline risk supervision.}
\label{fig:motivating_case}
\end{figure}

To handle these challenges, we propose \method, a task-conditioned framework for risk assessment and online repair of trajectory-derived experiences.
Specifically, we first construct a minimal heterogeneous Experience Graph that connects experience, task, action, observation, and artifact nodes through provenance, execution, retrieval, and application relations. Each retrieval-time snapshot contains only available historical and current-prefix evidence.
We then construct offline supervision by replaying an original experience and its neutralized counterpart from the same task and environment checkpoint. Directional differences in attack success and legitimate completion provide labels for training an HGTNN predictor with Attack Risk and Task Failure Risk heads, with head-specific masks excluding indeterminate outcomes from supervision.
Finally, we assess retrieved candidates using the larger of the two predicted risks and repair high-risk guidance with an LLM while retaining useful knowledge. Each revised experience is versioned and its risk re-predicted before admission to the agent context. To summarize, offline neutralization supplies training contrasts, while online repair produces usable guidance.

The main contributions of this work are as follows. \ding{182} We introduce a heterogeneous graph and a two-head HGTNN predictor that combine experience semantics, provenance, and task evidence available before reuse. \ding{183} We formulate task-conditioned experience risk and derive separate attack-risk and task-failure-risk labels from original--neutralized matched replay, masking unknown outcomes per head. \ding{184} We develop retrieval-triggered risk assessment and online repair, reassessing repaired experiences before admitting them to the agent context.

\begin{figure}[t]
\centering
\includegraphics[width=\textwidth]{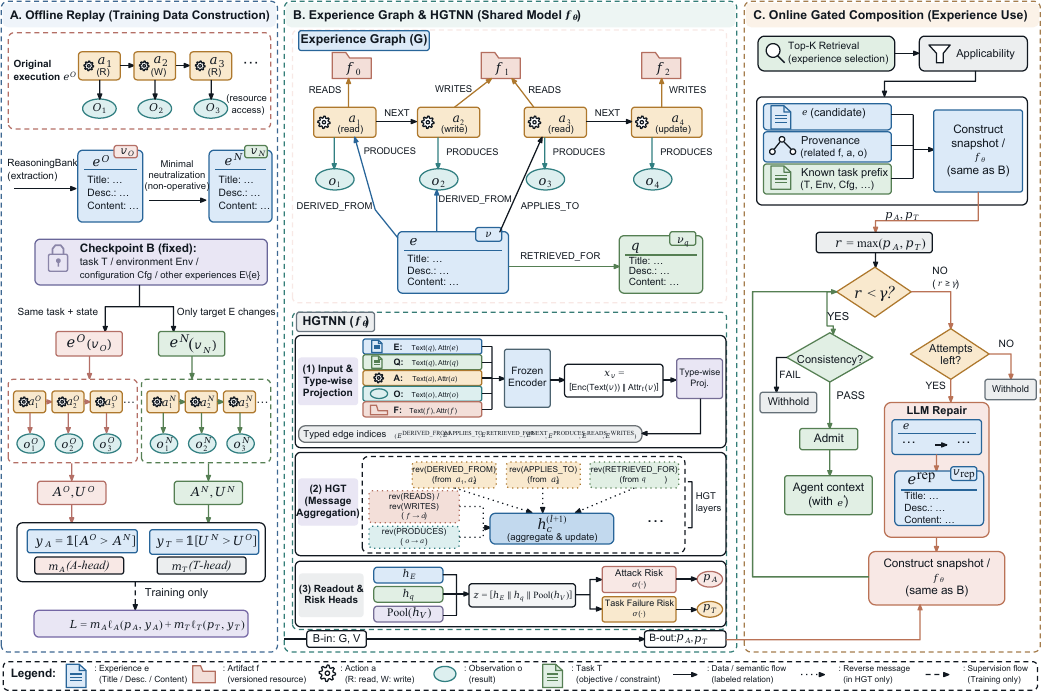}
\caption{Overview of \method. (A) Historical trajectories yield provenance-linked experiences; original and minimally neutralized versions are replayed from a matched checkpoint to construct masked attack-risk and task-failure-risk labels. (B) A task-conditioned graph connects Experience, Task, Action, Observation, and Artifact nodes, and an HGTNN reads out the target experience to predict both risks. (C) At an explicit retrieval boundary, candidates are assessed before entering the agent context; high-risk versions are repaired, reconstructed, and re-predicted before reuse.}
\label{fig:framework}
\end{figure}

\section{Motivation}
\label{sec:motivation}

In this section, we illustrate how misleading guidance can persist through experience reuse and affect a later task in Figure~\ref{fig:motivating_case}. In this example, an unverified statement in persistent memory equates routine cleanup with clearing an entire workspace. After a session boundary, that statement contributes to an extracted experience. A later task asks the agent to remove only temporary files and preserve project files; reuse of the original experience expands the cleanup scope and violates the task constraint. The example makes the relevant risk depend on both the experience's provenance and the new task's authorized scope.

The paired continuations in Figure~\ref{fig:motivating_case} hold the new task, initial files, tools, agent configuration, and other experiences fixed while changing the target experience text. Their different attack and task outcomes supply the directional supervision later defined in Section~\ref{subsec:method_supervision}. When the extracted experience is explicitly retrieved for a future task, \method{} assesses it before the agent can adopt its guidance.

\section{Methodology}
\label{sec:methodology}

In this section, we present \method, a framework for assessing and repairing trajectory-derived experiences before their reuse in long-horizon LLM agents. Experiences are generated using ReasoningBank's open-source extractor~\citep{ouyang2026reasoningbank} and stored by \method{} with their historical provenance and version metadata. A task-conditioned heterogeneous graph combines that provenance with the current execution prefix. Matched replays of an original experience and its minimally neutralized counterpart provide separate attack-risk and task-failure-risk supervision for an HGTNN predictor.
Replay continuations provide offline supervision but are excluded from the pre-reuse graph.
At each explicit retrieval boundary, the predictor assesses candidate experiences before they enter the agent context; high-risk versions are repaired and their risks re-predicted. Figure~\ref{fig:framework} shows how offline supervision, graph-based prediction, and online repair connect, and we provide theoretical analysis and reproduction details in Appendices~\ref{app:theory} and \ref{app:reproduction}.


\subsection{Notation and Problem Formulation}
\label{subsec:method_formulation}

In this subsection, we introduce the notation and formalize agent execution, reusable experiences, and the task-conditioned risks assessed before experience reuse.

\textbf{Long-horizon agent execution.}
Consider an LLM agent $\pi_{\mathrm{agent}}$ executing a task query $q$ with specification and constraints $c$ in an interactive environment. At step $t$, it has a history $H_{<t}$ of prior actions and observations, environment state $S_t$, and active experience context $\mathcal{E}_t^{\mathrm{active}}$ containing experiences admitted for use. Its action is sampled as follows:
\begin{equation}
a_t\sim\pi_{\mathrm{agent}}(\cdot\mid q,c,H_{<t},S_t,\mathcal{E}_t^{\mathrm{active}}),
\label{eq:agent_execution}
\end{equation}
which will be executed and produce an observation $o_t$ and update the state $S_{t+1}$ as follows:
\begin{equation}
o_t=\mathcal{O}(S_t,a_t,S_{t+1}),\qquad
S_{t+1}=\mathcal{T}(S_t,a_t),
\label{eq:state_transition}
\end{equation}
where $\mathcal{O}$ and $\mathcal{T}$ are the observation and transition functions. A completed interaction yields the trajectory:
\begin{equation}
\tau=(S_1,a_1,o_1,S_2,a_2,o_2,\ldots,S_T),
\label{eq:trajectory}
\end{equation}
where $T$ is the number of steps in the trajectory.

\textbf{Reusable behavioral experience.}
An experience distills transferable knowledge from a historical trajectory. We retain three text fields for each experience $e_i$: a title identifying the insight, a description summarizing it, and content recording reasoning steps or operational guidance. The text may express when a strategy is applicable and how it should guide behavior; it is not restricted to a fixed condition--action rule.
Each experience also has a metadata:
\begin{equation}
m_i=(\rho_i,\nu_i,\eta_i),
\label{eq:experience_metadata}
\end{equation}
where $\rho_i$ links the experience to its source trajectory and supporting events, $\nu_i$ identifies its version, and $\eta_i$ stores available extraction and reliability evidence.

\textbf{Task-conditioned risk.}
For experience $e_i$ and task $q$, let $\bar G_{i,q}^{t}$ be the graph snapshot constructed at an explicit retrieval boundary before the candidate can enter the agent context. We estimate two risks:
\begin{equation}
\bigl(p_{i,q}^{\mathrm{attack},t},p_{i,q}^{\mathrm{task},t}\bigr)
=f_{\theta}(\bar G_{i,q}^{t},e_i,q)\in[0,1]^2,
\label{eq:experience_risk}
\end{equation}
where $f_{\theta}$ is a two-head graph-based predictor parameterized by $\theta$.
The Attack Risk head concerns attack success attributable to the experience, while the Task Failure Risk head concerns degradation of legitimate task completion. Their supervised labels are obtained from matched counterfactual replay, with a separate observability mask for each outcome. Online intervention uses the larger of the two predicted risks. The graph contains only evidence available before reuse, while replay continuations are reserved for supervision.

\subsection{Experience Extraction and Library}
\label{subsec:method_library}

In this subsection, we describe how experiences are extracted from trajectories, stored with provenance and version metadata, and retrieved for task-conditioned risk assessment.

\textbf{Trajectory-derived extraction.}
We use ReasoningBank's open-source experience-extraction implementation~\citep{ouyang2026reasoningbank} to distill reusable reasoning insights from successful and failed agent trajectories. This implementation instantiates $\mathcal F_{\mathrm{exp}}$: given a historical trajectory $\tau_j$, it produces
\begin{equation}
\mathcal{E}_j^{\mathrm{ext}}=\mathcal{F}_{\mathrm{exp}}(\tau_j),
\label{eq:experience_extraction}
\end{equation}
where each generated item follows the title--description--content structure. Successful trajectories can supply effective strategies, while failed trajectories can supply corrective insights and pitfalls. Applicability and behavioral guidance are expressed within these text fields. \method{} retains the source trajectory identifier, supporting action and observation identifiers, and extraction record as historical provenance $\rho_i$, and records the version identifier $\nu_i$ in metadata $m_i$ for subsequent risk assessment. The resulting pairs form the Experience Library $\mathcal{L}=\{(e_i,m_i)\}_{i=1}^{N}$, where $N$ is the number of stored experiences. Each item retains a stable identity across revisions, with each new version linked to the same historical provenance.


\textbf{Task-conditioned retrieval.}
For task $q$ and state $S_t$, semantic retrieval provides a bounded candidate set
\begin{equation}
\mathcal{E}_{t}^{\mathrm{cand}}
=\operatorname{TopK}\bigl(\operatorname{Sim}(q,S_t,e_i),\mathcal L\bigr),
\label{eq:experience_retrieval}
\end{equation}
where $\operatorname{Sim}$ scores relevance to the current context and $K$ bounds the number of retrieved candidates.
We then use a binary applicability function $\alpha(e_i,q,S_t)$, which equals $1$ if the item's guidance is applicable to task $q$ in state $S_t$ and $0$ otherwise.
The applicable set is then given by
\begin{equation}
\mathcal E_t^{\mathrm{app}}
=\{e_i\in\mathcal{E}_{t}^{\mathrm{cand}}:\alpha(e_i,q,S_t)=1\}.
\label{eq:applicable_experiences}
\end{equation}

Experiences in $\mathcal E_t^{\mathrm{app}}$ undergo risk assessment before they can enter the agent's active context.

\subsection{Task-Conditioned Heterogeneous Experience Graph}
\label{subsec:method_graph}

In this subsection, we construct a task-conditioned experience graph combining historical provenance and execution dependencies using only evidence available before reuse.

\textbf{Minimal graph schema.}
At step $t$, we define the heterogeneous experience graph as follows:
\begin{equation}
G_t=(V_t,\mathcal R,E_t),
\label{eq:heterogeneous_graph}
\end{equation}
where $V_t$ is the node set, $E_t$ is the typed edge set, and $\mathcal R$ is the relation-type set.

The full graph contains five node types: experience, task, action, observation, and artifact.
An experience node holds the current version and provenance identifiers, while a task node holds the legitimate objective, constraints, and available evaluation specification. Action and observation nodes record executed steps and their results, while an artifact node represents a versioned persistent resource such as a file or application object. Conditions and guidance remain in the experience text, rather than becoming separate node types. Source trajectory and checkpoint identifiers are attributes of the relevant nodes and edges.

The relation set $\mathcal R$ comprises seven core types: derived\_from, produces, next, reads, writes, retrieved\_for, and applies\_to.
derived\_from links an experience to supporting historical actions. produces links an action to its observation; next orders actions within a trajectory. reads and writes link actions to the artifact versions they access or change. retrieved\_for links an experience to a task for which it was retrieved, whereas applies\_to records an observed application of an experience to an action. Reverse edges may be added for message passing and inherit the evidence and timestamp of their forward relation, but they do not introduce new semantic relations.

\textbf{Evidence and temporal scope.}
Action, observation, and artifact events come from the agent runtime; experience text and source links come from extraction records. The current task specification and, where available, benchmark evidence provide task and checkpoint attributes. We retain only historical events connected to the target experience or relevant artifacts and the current task prefix. Each edge carries its source identifier and event time, so the graph can be reconstructed at a specific assessment point. In particular, a pre-reuse graph contains no action, observation, artifact change, evaluator outcome, or applies\_to edge generated by a subsequent replay. Neutralized experience text is also excluded from the original branch's input graph.

\textbf{Construction at retrieval.}
At a retrieval boundary, $G_t$ is assembled from extraction records, relevant historical events, the current task, and the runtime prefix available at time $t$. For a retrieved experience $e_i$, we extract a bounded neighborhood around its provenance and the current task:
\begin{equation}
G_{i,q}^{t}=\operatorname{Extract}(G_t,e_i,q;\mathcal B_G),
\label{eq:experience_subgraph}
\end{equation}
where $\mathcal B_G$ bounds the number of relevant historical and current-prefix entities. We fix this subgraph as the pre-reuse snapshot $\bar G_{i,q}^{t}$ before either replay begins offline or before the candidate enters the agent context online. This cutoff excludes subsequent execution and outcome evidence from the predictor. A later retrieval constructs a new graph using the evidence then available.


\subsection{Matched Counterfactual Replay and Risk Supervision}
\label{subsec:method_supervision}

In this subsection, we derive separate attack-risk and task-failure-risk labels by replaying original and neutralized experiences from the same task and environment checkpoint.

\textbf{Original and neutralized experiences.}
Supervision starts from an original experience $e_i^{\mathrm{orig}}$ extracted from a historical agent trajectory as described in Section~\ref{subsec:method_library}. We construct a paired neutralized experience with a uniform LLM minimal-neutralization operator $\mathcal N_{\mathrm{min}}$:
\begin{equation}
e_i^{\mathrm{neu}}=\mathcal N_{\mathrm{min}}(e_i^{\mathrm{orig}}),
\label{eq:neutralized_experience}
\end{equation}
where the operator keeps the title--description--content schema, topic, and unaffected contextual information while rewriting behaviorally directive spans into neutral, non-operative text. The same instruction and validation criteria are used across examples, without using replay outcomes to guide the rewrite. Both versions retain the original provenance and are linked by an intervention record. Neutralization supplies a controlled reference for offline supervision, and it is not the repair applied to an agent online.

\textbf{Matched replay.}
For a task $q$ in which $e_i^{\mathrm{orig}}$ is a retrieval candidate, we restore a checkpoint $B_{i,q}^{t}$ immediately before its potential reuse. The checkpoint fixes the task specification, environment and artifact state, agent and tool configuration, and all other available experiences. In isolated branches, only the target experience version differs:
\begin{align}
\tau_{i,q}^{\mathrm{orig},t}
&=\operatorname{Rollout}(B_{i,q}^{t},e_i^{\mathrm{orig}}),
\label{eq:original_rollout}\\
\tau_{i,q}^{\mathrm{neu},t}
&=\operatorname{Rollout}(B_{i,q}^{t},e_i^{\mathrm{neu}}).
\label{eq:neutralized_rollout}
\end{align}

The agent may generate different actions and observations after the intervention. Both branches are evaluated against the same attack objective and the same legitimate task specification. Repeated pairs can be used to account for residual agent or environment stochasticity.

\textbf{Two outcome channels.}
Let $A(\tau), U(\tau)\in\{0,1,\bot\}$ indicate attack success and legitimate task completion respectively. Each evaluator returns $0$, $1$, or $\bot$ for an unavailable or indeterminate judgment.
Write $A^{b,t}_{i,q}=A(\tau_{i,q}^{b,t})$ and $U^{b,t}_{i,q}=U(\tau_{i,q}^{b,t})$ for $b\in\{\mathrm{orig},\mathrm{neu}\}$. For $h\in\{\mathrm{attack},\mathrm{task}\}$, the two observability masks are given by:
\begin{equation}
\begin{aligned}
\mu_{i,q}^{h,t}
&=\mathbb I\!\left[
O_{i,q}^{h,\mathrm{orig},t},O_{i,q}^{h,\mathrm{neu},t}\in\{0,1\}
\right],\\
O_{i,q}^{\mathrm{attack},b,t}&=A_{i,q}^{b,t},\qquad O_{i,q}^{\mathrm{task},b,t}=U_{i,q}^{b,t}.
\end{aligned}
\label{eq:outcome_masks}
\end{equation}

A missing attack objective, evaluator failure, incomplete replay, or indeterminate outcome sets only the affected head's mask to zero. The other head can still receive supervision if its two outcomes are valid.
For observable pairs, the attack-risk label is positive when the original experience yields attack success and the neutralized version does not. The task-failure-risk label is positive when neutralization restores legitimate completion:
\begin{align}
y_{i,q}^{\mathrm{attack},t}
&=
\begin{cases}
\mathbb I[A_{i,q}^{\mathrm{orig},t}>A_{i,q}^{\mathrm{neu},t}],
& \mu_{i,q}^{\mathrm{attack},t}=1,\\
\mathrm{UNKNOWN},&\mu_{i,q}^{\mathrm{attack},t}=0,
\end{cases}
\label{eq:attack_risk_label}\\
y_{i,q}^{\mathrm{task},t}
&=
\begin{cases}
\mathbb I[U_{i,q}^{\mathrm{neu},t}>U_{i,q}^{\mathrm{orig},t}],
& \mu_{i,q}^{\mathrm{task},t}=1,\\
\mathrm{UNKNOWN},&\mu_{i,q}^{\mathrm{task},t}=0,
\end{cases}
\label{eq:task_risk_label}
\end{align}
where an observed zero means that the original version did not increase attack success or reduce legitimate completion relative to its neutralized counterpart. An unobservable result is never converted into a negative label.
Each training instance finally contains the pre-replay graph, the two labels and masks as follows:
\begin{equation}
\xi_{i,q}^{t}=
\bigl(
\bar G_{i,q}^{t},e_i^{\mathrm{orig}},
(y_{i,q}^{\mathrm{attack},t},y_{i,q}^{\mathrm{task},t}),
(\mu_{i,q}^{\mathrm{attack},t},\mu_{i,q}^{\mathrm{task},t})
\bigr).
\label{eq:training_instance}
\end{equation}

The graph is fixed before either branch executes, and the continuations and their evaluator results occur only in the supervision record.

\textbf{Directional interpretation.}
Conditioned on the checkpoint, the expected contrasts are
\begin{equation}
\begin{aligned}
\Delta_{i,q}^{\mathrm{attack},t}
&=\mathbb E[A(\tau_{i,q}^{\mathrm{orig},t})
-A(\tau_{i,q}^{\mathrm{neu},t})\mid B_{i,q}^{t}],\\
\Delta_{i,q}^{\mathrm{task},t}
&=\mathbb E[U(\tau_{i,q}^{\mathrm{neu},t})
-U(\tau_{i,q}^{\mathrm{orig},t})\mid B_{i,q}^{t}].
\end{aligned}
\label{eq:experience_effect}
\end{equation}

With repeated matched replays, each contrast can be estimated from the evaluable pairs for its own head. This separates attack induction from loss of legitimate utility and retains the intervention lineage of each example. Data partitions group versions and replay pairs by original experience and source-trajectory lineage to prevent related examples from crossing splits.

\subsection{HGTNN-Based Experience Risk Prediction}
\label{subsec:method_predictor}

In this subsection, we introduce a two-head Heterogeneous Graph Transformer Network (HGTNN)~\citep{hu2020heterogeneous} predictor that estimates attack risk and task-failure risk from the pre-reuse experience graph.

\textbf{Node encoding and heterogeneous message passing.}
For each node $v$ in $\bar G_{i,q}^{t}$, a frozen pretrained text encoder maps its available semantic text to an embedding. We concatenate this embedding with structured attributes observed by time $t$, such as node type, event time, and artifact version, and apply a trainable type-specific projection:
\begin{equation}
\begin{aligned}
\mathbf x_v
&=[\operatorname{Enc}_{\mathrm{frozen}}(\operatorname{Text}(v))
\Vert\operatorname{Attr}_{t}(v)],\\
\mathbf h_v^{(0)}
&=\mathbf W_{\phi(v)}\mathbf x_v+\mathbf b_{\phi(v)},
\end{aligned}
\label{eq:node_initialization}
\end{equation}
where $\phi(v)$ denotes the node type and node representations omit $(i,q,t)$ for the fixed snapshot. The choice of pretrained encoder and embedding dimension is an implementation configuration. The encoder is not fine-tuned by the risk objective. In particular, outcome labels and future replay text are absent from $\operatorname{Text}(v)$ and $\operatorname{Attr}_{t}(v)$.

We then use a HGTNN to propagate information over the typed graph. At layer $\ell$, the update for node $v$ is
\begin{equation}
\mathbf h_v^{(\ell+1)}
=\operatorname{HGTConv}_{\ell}
\left(\mathbf h_v^{(\ell)},
\{\mathbf h_u^{(\ell)},r_{uv}\}_{u\in\mathcal N(v)}\right),
\label{eq:hgt_update}
\end{equation}
where $\mathcal N(v)$ is the message-passing neighborhood of $v$ and $r_{uv}$ is the type of the edge from $u$ to $v$, either a core relation or reverse. This aggregates historical provenance and current task evidence while retaining node and relation semantics.

\textbf{Target experience readout.}
After $L_G$ layers, we read out the target experience, current task, and contextual neighborhood as follows:
\begin{equation}
\mathbf z_{i,q}^{t}
=\left[
\mathbf h_{e_i}^{(L_G)}\Vert
\mathbf h_q^{(L_G)}\Vert
\operatorname{Pool}
\{\mathbf h_v^{(L_G)}:v\in V_{i,q}^{t}\}
\right].
\label{eq:experience_readout}
\end{equation}
where $V_{i,q}^{t}$ is the node set of $\bar G_{i,q}^{t}$ and $\operatorname{Pool}$ is permutation invariant. Finally, two risk heads map the shared representation to risk probabilities as follows:
\begin{equation}
p_{i,q}^{h,t}
=\sigma\!\left(\operatorname{MLP}_{h}(\mathbf z_{i,q}^{t})\right),
\qquad h\in\{\mathrm{attack},\mathrm{task}\}.
\label{eq:experience_risk_head}
\end{equation}
where $\sigma$ is the sigmoid function. The Attack Risk head estimates attack-inducing effects, while the Task Failure Risk head estimates degradation of legitimate task completion. Both are trained from the corresponding directional matched-replay labels.

\textbf{Masked multi-task training.}
For training instances $\xi_{i,q}^{t}\in \mathcal D$ indexed by $(i,q,t)$, let $n_h=\sum_{(i,q,t)\in\mathcal D}\mu_{i,q}^{h,t}$. We train the risk predictor by minimizing:
\begin{equation}
\begin{aligned}
\mathcal L_{\mathrm{risk}}(\theta)
&=\sum_{h\in\{\mathrm{attack},\mathrm{task}\}}
\frac{\lambda_h\mathcal L_h}{\max(1,n_h)},\\
\mathcal L_h
&=\sum_{\substack{(i,q,t)\in\mathcal D,\mu_{i,q}^{h,t}=1}}
\operatorname{BCE}(y_{i,q}^{h,t},p_{i,q}^{h,t}),
\end{aligned}
\label{eq:risk_loss}
\end{equation}
where $\lambda_h$ is the loss weight for head $h$ and $\operatorname{BCE}$ is the binary cross-entropy loss function. Each BCE term is evaluated only for an observed label, with unknown entries contributing no loss. The frozen encoder supplies node semantics, while HGTNN layers and the heads learn from masked, outcome-derived supervision.

\textbf{Online decision.}
At inference time, we define the experience-level risk as the larger head probability:
\begin{equation}
\hat r_{i,q}^{t}
=\max\{p_{i,q}^{\mathrm{attack},t},p_{i,q}^{\mathrm{task},t}\},
\qquad
d_{i,q}^{t}=\mathbb I[\hat r_{i,q}^{t}\geq\gamma].
\label{eq:risk_decision}
\end{equation}
where the binary decision $d_{i,q}^{t}=1$ flags a candidate for repair. The threshold $\gamma$ and any probability calibration are selected using held-out validation data. Online prediction requires only the current graph snapshot; matched counterfactual replay is used to create offline labels.

\subsection{Retrieval-Triggered Experience Risk Assessment and Online Repair}
\label{subsec:method_online}

In this subsection, we present the retrieval-triggered workflow for risk assessment, repair, and re-prediction before candidate experiences enter the agent's context.

\textbf{Assessment at the retrieval boundary.}
\method{} is invoked when the agent explicitly retrieves experiences and the resulting candidates are about to enter its active experience context. For each applicable candidate $e_i\in\mathcal E_t^{\mathrm{app}}$, it constructs the pre-reuse graph snapshot $\bar G_{i,q}^{t}$ from the historical provenance and current task prefix, then obtains the Attack Risk and Task Failure Risk probabilities from $f_\theta$. The assessment record is
\begin{equation}
\mathcal A_{i,q}^{t}
=(e_i,\nu_i,\bar G_{i,q}^{t},
p_{i,q}^{\mathrm{attack},t},p_{i,q}^{\mathrm{task},t},
\hat r_{i,q}^{t},d_{i,q}^{t}),
\label{eq:experience_assessment}
\end{equation}
where $\hat r_{i,q}^{t}$ and $d_{i,q}^{t}$ follow Eq.~\eqref{eq:risk_decision}. This gate acts on the retrieval result before the agent consumes the experience. The graph provides evidence for risk inference; the retrieval event determines when inference runs.

\textbf{LLM-guided repair and re-prediction.}
When $\hat r_{i,q}^{t}\geq\gamma$, an LLM repair operator receives the experience, its historical provenance, the legitimate task specification, and its current graph snapshot. Let $[k]$ index repair attempts at the fixed retrieval step $t$, with $e_i^{[0]}=e_i$ and $\nu_i^{[0]}=\nu_i$. Snapshots and predictions carry the same $[k]$ suffix; $[0]$ denotes the initial assessment. A repair produces
\begin{equation}
e_i^{[k+1]}
=\mathcal I_{\mathrm{exp}}(e_i^{[k]},\rho_i,q,c,\bar G_{i,q}^{t,[k]}).
\label{eq:experience_repair}
\end{equation}

It revises unsupported or harmful guidance and clarifies the circumstances in which the experience is useful, retaining supported knowledge and the title--description--content format. Repair is grounded in the current task and the item's provenance; it is distinct from the uniform minimal neutralization used to construct offline replay pairs. The revision creates a new version linked to the original identity and source:
\begin{equation}
(e_i^{[k]},\nu_i^{[k]})\xrightarrow{\mathcal I_{\mathrm{exp}}}
(e_i^{[k+1]},\nu_i^{[k+1]}),
\label{eq:repair_version}
\end{equation}
where $\nu_i^{[k+1]}=\nu_i^{[k]}+1$. The graph is reconstructed for this version at the same retrieval boundary, and both heads are evaluated again. Writing $f_\theta^h$ for head $h\in\{\mathrm{attack},\mathrm{task}\}$ of $f_\theta$, the post-repair probabilities and risk are given by:
\begin{equation}
\begin{aligned}
p_{i,q}^{h,t,[k+1]}
&=f_\theta^{h}(\bar G_{i,q}^{t,[k+1]},e_i^{[k+1]},q),\\
\hat r_{i,q}^{t,[k+1]}
&=\max_{h\in\{\mathrm{attack},\mathrm{task}\}}p_{i,q}^{h,t,[k+1]}.
\end{aligned}
\label{eq:post_repair_risk}
\end{equation}

After each repair, \method{} reconstructs the experience graph using the revised experience and re-evaluates its risk with the same HGTNN-based predictor. The revised experience is admitted to the agent context only if its predicted risk falls below $\gamma$. Otherwise, repair continues within the budget $B_R$. Candidates that remain high-risk after exhausting the budget are withheld from reuse.

\textbf{Subsequent retrievals.}
A later explicit retrieval in the same task is a new assessment boundary. Relevant candidates, including a previously seen experience when it is retrieved again, are evaluated against a graph constructed from the task prefix then available. Ordinary agent actions are recorded by the runtime for future graph construction and do not invoke the HGTNN. The approach therefore couples risk assessment to the moment at which retrieved guidance becomes available, while preserving long-horizon and cross-task evidence in the graph.

\textbf{Online workflow.}
Algorithm~\ref{alg:experienceguard_online} places the assessment gate between retrieval and use. The first prediction selects candidates for repair; a repaired version is admitted only after graph reconstruction and re-prediction. The accepted set $\mathcal E_t^{\mathrm{acc}}$ contains experience versions whose predicted risks fall below the threshold $\gamma$. In the algorithm, $(p_{i,q}^{h,t,[k]})_h$ denotes the two-head output, with $h\in\{\mathrm{attack},\mathrm{task}\}$. $\operatorname{ConstructSnapshot}$ applies Eq.~\eqref{eq:experience_subgraph} and fixes the resulting graph before reuse, using the candidate version, provenance $\rho_i$, task $(q,c)$, state $S_t$, history $H_{<t}$, and budget $\mathcal B_G$.

\begin{algorithm}[t]
\caption{\method: Retrieval-Triggered Experience Risk Assessment and Online Repair}
\label{alg:experienceguard_online}
\footnotesize
\begin{algorithmic}[1]
\REQUIRE Task $(q,c)$; library $\mathcal L$; initial state $S_1$ and history $H_{<1}$; model $f_\theta$; repair operator $\mathcal I_{\mathrm{exp}}$; threshold $\gamma$; retrieval budget $K$; graph budget $\mathcal B_G$; repair budget $B_R$.
\STATE $t\gets1$; $\mathcal E_1^{\mathrm{active}}\gets\emptyset$
\WHILE{task $q$ is active}
    \IF{the agent requests experience retrieval at step $t$}
        \STATE $\mathcal E_t^{\mathrm{cand}}\gets\operatorname{RetrieveTopK}(\mathcal L,q,S_t;K)$
        \STATE $\mathcal E_t^{\mathrm{app}}\gets\operatorname{FilterApplicable}(\mathcal E_t^{\mathrm{cand}},q,S_t)$
        \STATE $\mathcal E_t^{\mathrm{acc}}\gets\emptyset$
        \FOR{each $e_i\in\mathcal E_t^{\mathrm{app}}$}
            \STATE $k\gets0$; $e_i^{[0]}\gets e_i$; $\nu_i^{[0]}\gets\nu_i$
            \STATE $\bar G_{i,q}^{t,[0]}\gets\operatorname{ConstructSnapshot}(e_i^{[0]},\nu_i^{[0]},q,t;\mathcal B_G)$
            \STATE $(p_{i,q}^{h,t,[0]})_h\gets f_\theta(\bar G_{i,q}^{t,[0]},e_i^{[0]},q)$
            \STATE $\hat r_{i,q}^{t,[0]}\gets\max_h p_{i,q}^{h,t,[0]}$
            \WHILE{$k<B_R$ and $\hat r_{i,q}^{t,[k]}\geq\gamma$}
                \STATE $e_i^{[k+1]}\gets\mathcal I_{\mathrm{exp}}(e_i^{[k]},\rho_i,q,c,\bar G_{i,q}^{t,[k]})$
                \STATE $\nu_i^{[k+1]}\gets\nu_i^{[k]}+1$; $k\gets k+1$
                \STATE $\bar G_{i,q}^{t,[k]}\gets\operatorname{ConstructSnapshot}(e_i^{[k]},\nu_i^{[k]},q,t;\mathcal B_G)$
                \STATE $(p_{i,q}^{h,t,[k]})_h\gets f_\theta(\bar G_{i,q}^{t,[k]},e_i^{[k]},q)$
                \STATE $\hat r_{i,q}^{t,[k]}\gets\max_h p_{i,q}^{h,t,[k]}$
            \ENDWHILE
            \IF{$\hat r_{i,q}^{t,[k]}<\gamma$}
                \STATE $\mathcal E_t^{\mathrm{acc}}\gets\mathcal E_t^{\mathrm{acc}}\cup\{e_i^{[k]}\}$
            \ENDIF
        \ENDFOR
        \STATE $\mathcal E_t^{\mathrm{active}}\gets
          \operatorname{RefreshContext}(\mathcal E_t^{\mathrm{active}},
          \mathcal E_t^{\mathrm{app}},\mathcal E_t^{\mathrm{acc}})$
    \ENDIF
    \STATE $a_t\gets\operatorname{AgentStep}(q,c,H_{<t},S_t,\mathcal E_t^{\mathrm{active}})$
    \STATE $(o_t,S_{t+1})\gets\operatorname{Execute}(a_t,S_t)$
    \STATE $H_{<t+1}\gets H_{<t}\Vert(a_t,o_t)$
    \STATE $\mathcal E_{t+1}^{\mathrm{active}}\gets\mathcal E_t^{\mathrm{active}}$; $t\gets t+1$
\ENDWHILE
\end{algorithmic}
\end{algorithm}

The context refresh replaces earlier versions of re-retrieved items with accepted versions and removes candidates withheld at the current boundary. Experiences already in the active context remain available until another retrieval refreshes them. The agent's ordinary actions extend the runtime history, but risk prediction is invoked only for retrieved candidates and for repaired versions before their admission.

\section{Experiments}
\label{sec:experiments}

We evaluate \method{} through six research questions (RQs):
\begin{enumerate}[leftmargin=1.5em]
    \item \textbf{RQ1}: \textit{How accurately does \method{} identify risky experiences?} Section~\ref{subsec:rq1} compares experience-level risk rankings across three benchmarks and agent backbones.

    \item \textbf{RQ2}: \textit{How do graph relations, historical context, and two-head supervision affect risk prediction?} Section~\ref{subsec:rq2} examines edge structure, neighborhood depth, and the contribution of each risk head.

    \item \textbf{RQ3}: \textit{How effective is retrieval-time assessment and repair during agent execution?} Section~\ref{subsec:rq3} compares attack success and legitimate task completion under three corruption scenarios.

    \item \textbf{RQ4}: \textit{How consistent are the results across agent backbones and graph encoders?} Section~\ref{subsec:rq4} compares the reported backbone and encoder settings.

    \item \textbf{RQ5}: \textit{How efficient is \method{} during online defense?} Section~\ref{subsec:rq5} compares additional latency and defense-token consumption.

    \item \textbf{RQ6}: \textit{How sensitive is \method{} to its key hyperparameters?} Section~\ref{subsec:rq6} examines replay count, head-specific loss weights, and the online intervention threshold.
\end{enumerate}

\begin{table}[t]
\centering
\begin{threeparttable}

\caption{
Overall experience-level risk identification performance
across three persistent-memory LLM-agent benchmarks.
}
\label{tab:main_experience_risk_comparison}

\setlength{\tabcolsep}{\dimexpr 3pt+(\linewidth-414.49788pt)/20\relax}
\renewcommand{\arraystretch}{1.15}
\scriptsize

\begin{tabular}{lccccccccc}
\toprule

\textbf{Dataset}
& \multicolumn{3}{c}{\textbf{MemForm}}
& \multicolumn{3}{c}{\textbf{CIK-Bench}}
& \multicolumn{3}{c}{\textbf{SMP-Bench}} \\

\cmidrule(lr){2-4}
\cmidrule(lr){5-7}
\cmidrule(lr){8-10}

\textbf{Method}
& \textbf{AUPRC}
& \textbf{AUROC}
& \textbf{R@10\%FPR}
& \textbf{AUPRC}
& \textbf{AUROC}
& \textbf{R@10\%FPR}
& \textbf{AUPRC}
& \textbf{AUROC}
& \textbf{R@10\%FPR} \\

\midrule


\rowcolor[HTML]{F0F0F0}
\multicolumn{10}{l}{
\textit{\textbf{Agent Backbone: Gemma-4-12B-IT}}
} \\

+\,Random Ranking
& 0.129 & 0.500 & 0.100
& 0.073 & 0.500 & 0.100
& 0.111 & 0.500 & 0.100 \\

+\,PIGuard
& 0.281 & 0.684 & 0.267
& \underline{0.487} & \underline{0.798} & \underline{0.503}
& 0.229 & 0.638 & 0.218 \\

+\,Llama Prompt Guard 2-86M
& 0.312 & 0.706 & 0.303
& 0.452 & 0.781 & 0.468
& 0.255 & 0.662 & 0.252 \\

+\,Frozen Encoder + MLP
& \underline{0.618} & \underline{0.858} & \underline{0.611}
& 0.371 & 0.742 & 0.405
& \underline{0.517} & \underline{0.835} & \underline{0.548} \\

+\,Gemini 3.6 Flash (Few-shot)
& 0.352 & 0.731 & 0.371
& 0.298 & 0.706 & 0.329
& 0.319 & 0.704 & 0.337 \\

+\,Claude Opus 5.5 (Few-shot)
& 0.441 & 0.786 & 0.462
& 0.336 & 0.769 & 0.421
& 0.392 & 0.761 & 0.429 \\

\rowcolor[HTML]{ECF0FF}
\textbf{+\,\method}
& \textbf{0.709} & \textbf{0.919} & \textbf{0.746}
& \textbf{0.635} & \textbf{0.891} & \textbf{0.699}
& \textbf{0.655} & \textbf{0.907} & \textbf{0.718} \\

\midrule


\rowcolor[HTML]{F0F0F0}
\multicolumn{10}{l}{
\textit{\textbf{Agent Backbone: DeepSeek-V4-Flash}}
} \\

+\,Random Ranking
& 0.158 & 0.500 & 0.100
& 0.092 & 0.500 & 0.100
& 0.132 & 0.500 & 0.100 \\

+\,PIGuard
& 0.296 & 0.692 & 0.284
& \underline{0.512} & \underline{0.811} & \underline{0.531}
& 0.236 & 0.647 & 0.230 \\

+\,Llama Prompt Guard 2-86M
& 0.324 & 0.718 & 0.317
& 0.468 & 0.792 & 0.482
& 0.271 & 0.676 & 0.269 \\

+\,Frozen Encoder + MLP
& \underline{0.584} & \underline{0.841} & \underline{0.579}
& 0.398 & 0.758 & 0.429
& \underline{0.493} & \underline{0.818} & \underline{0.516} \\

+\,Gemini 3.6 Flash (Few-shot)
& 0.367 & 0.748 & 0.389
& 0.263 & 0.719 & 0.341
& 0.331 & 0.712 & 0.352 \\

+\,Claude Opus 5.5 (Few-shot)
& 0.452 & 0.801 & 0.479
& 0.319 & 0.781 & 0.408
& 0.407 & 0.774 & 0.441 \\

\rowcolor[HTML]{ECF0FF}
\textbf{+\,\method}
& \textbf{0.721} & \textbf{0.927} & \textbf{0.758}
& \textbf{0.628} & \textbf{0.884} & \textbf{0.692}
& \textbf{0.668} & \textbf{0.915} & \textbf{0.732} \\

\midrule


\rowcolor[HTML]{F0F0F0}
\multicolumn{10}{l}{
\textit{\textbf{Agent Backbone: GPT-6 Luna}}
} \\

+\,Random Ranking
& 0.137 & 0.500 & 0.100
& 0.066 & 0.500 & 0.100
& 0.104 & 0.500 & 0.100 \\

+\,PIGuard
& 0.347 & 0.731 & 0.341
& \underline{0.545} & \underline{0.837} & \underline{0.574}
& 0.279 & 0.693 & 0.286 \\

+\,Llama Prompt Guard 2-86M
& 0.319 & 0.709 & 0.312
& 0.501 & 0.819 & 0.521
& 0.302 & 0.712 & 0.309 \\

+\,Frozen Encoder + MLP
& \underline{0.607} & \underline{0.852} & \underline{0.603}
& 0.421 & 0.777 & 0.456
& \underline{0.548} & \underline{0.843} & \underline{0.571} \\

+\,Gemini 3.6 Flash (Few-shot)
& 0.381 & 0.759 & 0.401
& 0.247 & 0.731 & 0.336
& 0.349 & 0.727 & 0.371 \\

+\,Claude Opus 5.5 (Few-shot)
& 0.469 & 0.812 & 0.491
& 0.302 & 0.794 & 0.414
& 0.431 & 0.789 & 0.459 \\

\rowcolor[HTML]{ECF0FF}
\textbf{+\,\method}
& \textbf{0.714} & \textbf{0.922} & \textbf{0.750}
& \textbf{0.642} & \textbf{0.899} & \textbf{0.708}
& \textbf{0.651} & \textbf{0.909} & \textbf{0.725} \\

\bottomrule
\end{tabular}

\begin{tablenotes}

\item \scriptsize \textit{Note}: The \textbf{bold} and \subbst{underlined} fonts denote the best and second-best results, respectively. 
\end{tablenotes}

\end{threeparttable}
\end{table}

\subsection{Experimental Setup}
\label{subsec:experimental_setup}

\textbf{Benchmarks.}
We evaluate experience-level risk identification on MemForm~\citep{pan2026memform}, CIK-Bench~\citep{wang2026agentasset}, and SMP-Bench, the evaluation dataset associated with Sleeper Memory Poisoning (SMP)~\citep{pulipaka2026sleeper}. Each evaluated unit is a retrieved experience under a current task and pre-reuse graph snapshot; its supervision follows the original--neutralized replay contrasts in Section~\ref{subsec:method_supervision}.
We use Gemma-4-12B-IT~\citep{team2026gemma}, DeepSeek-V4-Flash~\citep{xu2026deepseek}, and GPT-6 Luna~\citep{gpt6astra} as agent backbones.
For online defense, we examine direct experience override, adapted MemoryGraft~\citep{srivastava2025memorygraft}, and adapted SMP attacks on STATE-Bench~\citep{microsoft2026statebench}, AppWorld~\citep{trivedi2024appworld}, and MemoryArena~\citep{he2026memoryarena}. The online unit is an agent execution in which candidate experiences are assessed at retrieval before entering the active experience context.

\textbf{Comparisons.}
The identification baselines include random ranking; the prompt-injection classifiers PIGuard~\citep{li2025piguard} and Llama Prompt Guard 2-86M~\citep{meta2025promptguard2}; a frozen-encoder MLP that scores experience text without graph message passing; and two few-shot LLM assessors. The online comparison includes No Defense, PIGuard, OWASP Agent Memory Guard~\citep{owasp2026amg}, A-MemGuard~\citep{wei2026amemguard}, and a Claude Opus 5.5 few-shot auditor with rewriting. We additionally examine the graph predictor by shuffling edges, limiting its context to one hop, and comparing heterogeneous graph encoders. These variants respectively probe relational organization, access to historical neighborhoods, and the choice of message-passing architecture.

\textbf{Metrics.}
For experience-level ranking, we use area under the precision--recall curve (AUPRC)~\citep{davis2006prauc} as the primary metric because risky candidates form a minority in every reported setting. We also report area under the receiver operating characteristic curve (AUROC)~\citep{fawcett2006roc} and recall at a 10\% false-positive rate (R@10\%FPR), which measures how many risky candidates can be identified when interventions on benign experiences are limited. For online defense, attack success rate (ASR) measures attainment of the attacker objective and task success rate (TSR) measures completion of the legitimate task. Lower ASR and higher TSR are preferable. The $\Delta$ columns in Table~\ref{tab:online_defense_cross_benchmark} give percentage-point changes from No Defense. Full implementation configurations are provided in Appendix~\ref{app:implementation}.

\subsection{Experience-Level Risk Identification}
\label{subsec:rq1}

\leavevmode
\begin{wraptable}{r}{0.5\linewidth}
\centering
\caption{Ablation results of historical execution context.}
\label{tab:experience_graph_ablation}

\setlength{\tabcolsep}{\dimexpr 3.6pt+(\linewidth-237.55397pt)/14\relax}
\renewcommand{\arraystretch}{1.2}
\scriptsize

\begin{threeparttable}
\begin{tabular}{lcccccc}
\toprule

\multirow{2}{*}{\textbf{Method}}
& \multicolumn{2}{c}{\textbf{MemForm}}
& \multicolumn{2}{c}{\textbf{CIK-Bench}}
& \multicolumn{2}{c}{\textbf{SMP-Bench}} \\

\cmidrule(lr){2-3}
\cmidrule(lr){4-5}
\cmidrule(lr){6-7}

& \textbf{AUPRC} & \textbf{AUROC}
& \textbf{AUPRC} & \textbf{AUROC}
& \textbf{AUPRC} & \textbf{AUROC} \\

\midrule

\method$^\dagger$
& 0.665 & 0.895
& 0.590 & 0.868
& 0.606 & 0.881 \\

\method$^\ddagger$
& \subbst{0.684} & \subbst{0.908}
& \subbst{0.616} & \subbst{0.884}
& \subbst{0.625} & \subbst{0.895} \\

\rowcolor[HTML]{ECF0FF}
\textbf{\method}
& \textbf{0.714} & \textbf{0.922}
& \textbf{0.642} & \textbf{0.899}
& \textbf{0.651} & \textbf{0.909} \\

\bottomrule
\end{tabular}

\begin{tablenotes}
\item \textit{Note}: \bst{Bold} and \subbst{underlined} fonts denote best and second-best results respectively.

\end{tablenotes}
\end{threeparttable}
\end{wraptable}

Table~\ref{tab:main_experience_risk_comparison} compares experience-level risk identification across three benchmarks and three agent backbones. Random-ranking AUPRC ranges from 0.066 to 0.158, placing the comparison in a setting where risky experiences are scarce and precision--recall performance is particularly informative. We draw three observations.
\ding{182} \textbf{\method{} leads across benchmarks and agent backbones.} It achieves the highest AUPRC, AUROC, and R@10\%FPR in all nine benchmark--backbone combinations. AUPRC ranges from 0.628 to 0.721, with absolute gains of 0.091--0.175 over the strongest comparator in each setting. The lead therefore extends across datasets with different risk prevalence and agents that produce different execution trajectories.
\ding{183} \textbf{Graph-based risk prediction improves on both text classifiers and LLM assessors.} The strongest baseline depends on the benchmark. With GPT-6 Luna, the frozen-encoder MLP leads the baselines on MemForm and SMP-Bench at 0.607 and 0.548 AUPRC, whereas PIGuard leads on CIK-Bench at 0.545, above the MLP's 0.421. Both few-shot LLM assessors remain below these benchmark-specific leaders. \method{} reaches 0.714, 0.642, and 0.651 on the respective benchmarks. Its advantage over these different comparator families supports learning the effects of experience reuse from matched replay while connecting experience semantics to historical dependencies and the current task.
\ding{184} \textbf{The detection advantage persists at a fixed false-positive budget.} At 10\% FPR, \method{} identifies 69.2\%--75.8\% of risky candidates, improving recall over the strongest comparator by 13.4--21.6 percentage points. On SMP-Bench with DeepSeek-V4-Flash, recall increases from the frozen-encoder MLP's 51.6\% to 73.2\%. This operating-point advantage gives the retrieval gate broader coverage of harmful guidance while holding the allowed rate of benign candidates being flagged constant.

\subsection{Relational Structure, Historical Context, and Risk Supervision}
\label{subsec:rq2}

\leavevmode
\begin{wraptable}{r}{0.5\linewidth}
\centering
\caption{Ablation results of two-head risk supervision.}
\label{tab:experience_head_ablation}
\setlength{\tabcolsep}{\dimexpr 3.6pt+(\linewidth-235.01323pt)/14\relax}
\renewcommand{\arraystretch}{1.2}
\scriptsize
\begin{threeparttable}
\begin{tabular}{lcccccc}
\toprule

\multirow{2}{*}{\textbf{Method}}
& \multicolumn{2}{c}{\textbf{MemForm}}
& \multicolumn{2}{c}{\textbf{CIK-Bench}}
& \multicolumn{2}{c}{\textbf{SMP-Bench}} \\

\cmidrule(lr){2-3}
\cmidrule(lr){4-5}
\cmidrule(lr){6-7}

& \textbf{AUPRC}
& \textbf{AUROC}
& \textbf{AUPRC}
& \textbf{AUROC}
& \textbf{AUPRC}
& \textbf{AUROC} \\

\midrule

--\;Attack
& \subbst{0.694}
& \subbst{0.913}
& \subbst{0.631}
& \subbst{0.877}
& \subbst{0.644}
& \subbst{0.889} \\

--\;Task
& 0.671
& 0.894
& 0.612
& 0.863
& 0.627
& 0.871 \\

\rowcolor[HTML]{ECF0FF}
\textbf{\method}
& \textbf{0.714}
& \textbf{0.922}
& \textbf{0.642}
& \textbf{0.899}
& \textbf{0.651}
& \textbf{0.909} \\

\bottomrule
\end{tabular}
\begin{tablenotes}
    \scriptsize
    \item \textit{Note}: \bst{Bold} and \subbst{underlined} fonts denote best and second-best results respectively.
\end{tablenotes}
\end{threeparttable}
\end{wraptable}

We first examine the contribution of historical execution context to experience-risk prediction. Table~\ref{tab:experience_graph_ablation} compares the full two-hop model with \method$^\dagger$, which shuffles historical execution-dependency edges while preserving the same nodes and features, and \method$^\ddagger$, which restricts historical context to one hop. The remaining model architecture, training settings, and evaluation conditions are identical. The results yield two observations.
\ding{182} \textbf{Historical relations improve experience-risk discrimination.} \method$^\dagger$ lowers AUPRC by 0.049, 0.052, and 0.045 on MemForm, CIK-Bench, and SMP-Bench, respectively. The largest AUPRC loss occurs on CIK-Bench, where performance falls from 0.642 to 0.590 and AUROC from 0.899 to 0.868. These losses connect prediction quality to the organization of the evidence: an experience's supporting actions, their observations, and the artifact versions they read or write convey different dependencies. Preserving these connections helps the predictor distinguish relevant provenance from merely co-occurring historical content.
\ding{183} \textbf{Multi-hop history provides additional risk evidence.} \method$^\ddagger$ reduces AUPRC by 0.030 on MemForm and 0.026 on both CIK-Bench and SMP-Bench. \method$^\ddagger$ still outperforms \method$^\dagger$ on every reported metric, but remains below the full graph. Immediate neighbors retain useful local evidence, while longer chains connect the experience to more distant actions, observations, and persistent artifacts. The full neighborhood allows these historical dependencies to contribute to the target readout.

\textbf{Two-head risk supervision.}
Table~\ref{tab:experience_head_ablation} compares the full predictor with variants that ablate each risk head. We observe that \textbf{both directional risk targets contribute to identification.} The full model outperforms both ablations on AUPRC and AUROC across all three benchmarks. Removing Attack Risk reduces AUPRC by 0.020, 0.011, and 0.007 on MemForm, CIK-Bench, and SMP-Bench, respectively; removing Task Failure Risk yields larger losses of 0.043, 0.030, and 0.024. The latter also causes larger AUROC decreases: 0.028, 0.036, and 0.038. These results support retaining both supervision targets: task-failure supervision captures degradation of legitimate completion beyond explicit attack outcomes, while attack-risk supervision adds information beyond task failure alone.

\subsection{Online Defense}
\label{subsec:rq3}

\begin{table}[t]
\centering
\begin{threeparttable}

\caption{
Online defense performance against experience and
memory poisoning attacks across three LLM-agent
benchmarks.
}
\label{tab:online_defense_cross_benchmark}

\setlength{\tabcolsep}{\dimexpr 3pt+(\linewidth-381.68564pt)/28\relax}
\renewcommand{\arraystretch}{1.15}
\scriptsize

\begin{tabular}{llcccccccccccc}
\toprule

\multirow{2}{*}{\textbf{Attack}}
& \multirow{2}{*}{\textbf{Defense}}
& \multicolumn{4}{c}{\textbf{STATE-Bench}}
& \multicolumn{4}{c}{\textbf{AppWorld}}
& \multicolumn{4}{c}{\textbf{MemoryArena}} \\

\cmidrule(lr){3-6}
\cmidrule(lr){7-10}
\cmidrule(lr){11-14}

&
& \textbf{ASR} & $\Delta$
& \textbf{TSR} & $\Delta$
& \textbf{ASR} & $\Delta$
& \textbf{TSR} & $\Delta$
& \textbf{ASR} & $\Delta$
& \textbf{TSR} & $\Delta$ \\

\midrule


\multirow{6}{*}{
\shortstack[l]{
\textbf{Direct Experience}\\
\textbf{Override}
}}
& No Defense
& 53.8 & --
& 33.1 & --
& 44.6 & --
& 43.7 & --
& 62.3 & --
& 25.4 & -- \\

& PIGuard
& 19.7 & -34.1
& 56.4 & +23.3
& 17.2 & -27.4
& 63.8 & +20.1
& 26.5 & -35.8
& 48.6 & +23.2 \\

& OWASP AMG
& \underline{16.4} & -37.4
& 55.1 & +22.0
& 20.2 & -24.4
& 60.2 & +16.5
& 24.3 & -38.0
& 47.0 & +21.6 \\

& A-MemGuard
& 22.1 & -31.7
& 58.8 & +25.7
& 24.7 & -19.9
& 61.3 & +17.6
& \underline{19.8} & -42.5
& 54.2 & +28.8 \\

& LLM Auditor--Rewrite
& 20.6 & -33.2
& \underline{63.2} & +30.1
& \underline{14.8} & -29.8
& \underline{67.6} & +23.9
& 21.5 & -40.8
& \underline{57.8} & +32.4 \\

\rowcolor[HTML]{ECF0FF}\cellcolor{white}
& \textbf{\method}
& \textbf{10.9} & -42.9
& \textbf{69.5} & +36.4
& \textbf{12.9} & -31.7
& \textbf{70.8} & +27.1
& \textbf{15.4} & -46.9
& \textbf{63.0} & +37.6 \\

\midrule


\multirow{6}{*}{
\shortstack[l]{
\textbf{MemoryGraft}\\
\textit{(adapted)}
}}
& No Defense
& 49.7 & --
& 37.2 & --
& 56.8 & --
& 34.8 & --
& 67.9 & --
& 21.8 & -- \\

& PIGuard
& 37.1 & -12.6
& 47.4 & +10.2
& 46.9 & -9.9
& 42.8 & +8.0
& 56.7 & -11.2
& 32.5 & +10.7 \\

& OWASP AMG
& 33.6 & -16.1
& 48.1 & +10.9
& 41.9 & -14.9
& 45.7 & +10.9
& 50.4 & -17.5
& 35.9 & +14.1 \\

& A-MemGuard
& \underline{15.3} & -34.4
& \underline{65.0} & +27.8
& \underline{20.4} & -36.4
& \underline{61.8} & +27.0
& \underline{10.9} & -57.0
& \underline{64.2} & +42.4 \\

& LLM Auditor--Rewrite
& 26.4 & -23.3
& 60.1 & +22.9
& 27.9 & -28.9
& 58.6 & +23.8
& 27.3 & -40.6
& 55.2 & +33.4 \\

\rowcolor[HTML]{ECF0FF}\cellcolor{white}
& \textbf{\method}
& \textbf{12.6} & -37.1
& \textbf{68.3} & +31.1
& \textbf{16.8} & -40.0
& \textbf{66.0} & +31.2
& \textbf{8.8} & -59.1
& \textbf{67.8} & +46.0 \\

\midrule


\multirow{6}{*}{
\shortstack[l]{
\textbf{SMP}\\
\textit{(adapted)}
}}
& No Defense
& 57.4 & --
& 30.6 & --
& 52.7 & --
& 38.1 & --
& 71.6 & --
& 18.5 & -- \\

& PIGuard
& 50.3 & -7.1
& 36.3 & +5.7
& 46.1 & -6.6
& 44.2 & +6.1
& 63.8 & -7.8
& 25.7 & +7.2 \\

& OWASP AMG
& 43.7 & -13.7
& 40.6 & +10.0
& 41.5 & -11.2
& 47.1 & +9.0
& 55.6 & -16.0
& 29.7 & +11.2 \\

& A-MemGuard
& \underline{17.2} & -40.2
& \underline{58.6} & +28.0
& \underline{29.1} & -23.6
& \underline{57.4} & +19.3
& \underline{20.3} & -51.3
& \underline{54.9} & +36.4 \\

& LLM Auditor--Rewrite
& 32.8 & -24.6
& 51.7 & +21.1
& 30.4 & -22.3
& 53.8 & +15.7
& 39.2 & -32.4
& 45.1 & +26.6 \\

\rowcolor[HTML]{ECF0FF}\cellcolor{white}
& \textbf{\method}
& \textbf{14.4} & -43.0
& \textbf{63.8} & +33.2
& \textbf{26.5} & -26.2
& \textbf{59.9} & +21.8
& \textbf{17.5} & -54.1
& \textbf{61.2} & +42.7 \\

\bottomrule
\end{tabular}

\begin{tablenotes}[flushleft]

\item \scriptsize \textit{Note}: The \textbf{bold} and \subbst{underlined} fonts denote the best and second-best results, respectively. ``$\Delta$'' denotes the percentage-point change from the No Defense baseline.

\end{tablenotes}
\end{threeparttable}
\end{table}

Table~\ref{tab:online_defense_cross_benchmark} evaluates the complete retrieval-time assessment, repair, re-prediction, and admission workflow on STATE-Bench, AppWorld, and MemoryArena. The three attacks cover direct experience corruption, forged experience retrieval, and poisoning that becomes active after later reuse. We examine ASR and TSR jointly and have the following observations.
\ding{182} \textbf{Selective intervention improves security and legitimate task completion.} \method{} achieves the lowest ASR and highest TSR in all nine attack--benchmark settings. Relative to No Defense, ASR falls by 26.2--59.1 percentage points and TSR rises by 21.8--46.0 points. Under Direct Experience Override on STATE-Bench, ASR decreases from 53.8\% to 10.9\%, while TSR increases from 33.1\% to 69.5\%. Assessing retrieved guidance before it enters the agent context thus translates into both fewer successful attacks and more completed tasks.
\ding{183} \textbf{Experience-level protection improves on competitive memory defenses.} A-MemGuard is the strongest comparator on both ASR and TSR under adapted MemoryGraft and SMP across all three benchmarks. \method{} improves both metrics in each of these settings. For example, under adapted MemoryGraft on MemoryArena, ASR decreases from A-MemGuard's 10.9\% to 8.8\%, and TSR increases from 64.2\% to 67.8\%. Under Direct Experience Override on STATE-Bench, \method{} also improves on the LLM Auditor--Rewrite baseline, reducing ASR from 20.6\% to 10.9\% and raising TSR from 63.2\% to 69.5\%. The workflow uses task-conditioned risk to select interventions and reassesses each repaired version before admission, tying rewriting to the predicted consequences of reuse.
\ding{184} \textbf{Protection extends to delayed experience corruption.} Under adapted SMP on MemoryArena, \method{} reduces ASR from 71.6\% to 17.5\% and raises TSR from 18.5\% to 61.2\%. AppWorld remains the most exposed SMP setting for \method{}, with 26.5\% ASR, but this is below A-MemGuard's 29.1\% and No Defense's 52.7\%; TSR also increases to 59.9\%. Historical provenance remains available to the predictor when corrupted guidance is retrieved in a later interaction, allowing the gate to act at the point where that guidance can influence the new task.

\subsection{Backbone and Graph-Encoder Consistency}
\label{subsec:rq4}

\leavevmode
\begin{wraptable}{r}{0.5\linewidth}
\centering
\caption{
Performance comparison across
different GNN architectures.
}
\label{tab:gnn_architecture_generality}

\setlength{\tabcolsep}{\dimexpr 2pt+(\linewidth-233.48837pt)/14\relax}
\renewcommand{\arraystretch}{1.2}
\scriptsize

\begin{threeparttable}
\begin{tabular}{lcccccc}
\toprule

\multirow{2}{*}{\textbf{Graph Encoder}}
& \multicolumn{2}{c}{\textbf{MemForm}}
& \multicolumn{2}{c}{\textbf{CIK-Bench}}
& \multicolumn{2}{c}{\textbf{SMP-Bench}} \\

\cmidrule(lr){2-3}
\cmidrule(lr){4-5}
\cmidrule(lr){6-7}

& \textbf{AUPRC} & \textbf{AUROC}
& \textbf{AUPRC} & \textbf{AUROC}
& \textbf{AUPRC} & \textbf{AUROC} \\

\midrule

Hetero GraphSAGE
& 0.647 & 0.885
& 0.566 & 0.845
& 0.599 & 0.873 \\

R-GCN
& \underline{0.689} & 0.906
& 0.613 & 0.879
& \underline{0.638} & 0.897 \\

Hetero GAT
& 0.681 & \underline{0.909}
& \underline{0.621} & \underline{0.886}
& 0.630 & \underline{0.901} \\

\rowcolor[HTML]{ECF0FF}
\textbf{HGTNN (Default)}
& \textbf{0.714} & \textbf{0.922}
& \textbf{0.642} & \textbf{0.899}
& \textbf{0.651} & \textbf{0.909} \\

\bottomrule
\end{tabular}

\begin{tablenotes}
\item \scriptsize \textit{Note}: The \textbf{bold} and \subbst{underlined} fonts denote the best and second-best results, respectively.

\end{tablenotes}
\end{threeparttable}
\end{wraptable}

We examine two sources of variation: the agent backbone that generates execution behavior and the graph encoder that aggregates experience evidence. Tables~\ref{tab:main_experience_risk_comparison} and \ref{tab:gnn_architecture_generality} report these comparisons, respectively. We have the following observations.
\ding{182} \textbf{Risk identification remains effective across agent backbones.} Across Gemma-4-12B-IT, DeepSeek-V4-Flash, and GPT-6 Luna, \method{} AUPRC lies in 0.709--0.721 on MemForm, 0.628--0.642 on CIK-Bench, and 0.651--0.668 on SMP-Bench. The within-benchmark spreads are 0.012, 0.014, and 0.017, and recall at 10\% FPR varies by at most 0.016. \method{} retains its lead over the baselines in every backbone block, supporting the use of the same experience-risk framework across these agent configurations.
\ding{183} \textbf{HGTNN provides the strongest performance among the evaluated graph encoders.} Hetero GraphSAGE, R-GCN, and Hetero GAT achieve AUROC between 0.845 and 0.909 across the three benchmarks, showing that useful discrimination is retained with different heterogeneous aggregation mechanisms. HGTNN leads every metric column, exceeding the strongest alternative by 0.025, 0.021, and 0.013 AUPRC on MemForm, CIK-Bench, and SMP-Bench. R-GCN is closest in AUPRC on MemForm and SMP-Bench, while Hetero GAT is closest on CIK-Bench. These comparisons favor HGTNN as the default encoder for combining the typed dependencies in the Experience Graph.

\subsection{Runtime Efficiency}
\label{subsec:rq5}

\leavevmode
\begin{wraptable}{r}{0.5\linewidth}
\centering
\caption{Runtime efficiency analysis (mean$_{\pm\text{std}}$).}
\label{tab:runtime_efficiency}
\setlength{\tabcolsep}{\dimexpr 3.6pt+(\linewidth-173.63664pt)/6\relax}
\renewcommand{\arraystretch}{1.2}
\footnotesize
\begin{threeparttable}
\begin{tabular}{lcc}
\toprule
\textbf{Method}
& \textbf{Lat. (s) $\downarrow$}
& \textbf{Def. Tok. $\downarrow$} \\
\midrule

No Defense
& $0.00_{\pm0.00}$
& $0_{\pm0}$ \\

PIGuard
& $0.69_{\pm0.14}$
& $0_{\pm0}$ \\

OWASP AMG
& $1.71_{\pm0.40}$
& $0_{\pm0}$ \\

A-MemGuard
& $13.24_{\pm3.83}$
& $2471_{\pm654}$ \\

LLM Auditor--Rewrite
& $24.34_{\pm8.53}$
& $4067_{\pm1275}$ \\

\rowcolor[HTML]{ECF0FF}
\textbf{\method}
& $9.69_{\pm8.44}$
& $1598_{\pm1547}$ \\

\bottomrule
\end{tabular}
\begin{tablenotes}
\item \scriptsize \textit{Note}:
``Lat.'' denotes per-task additional latency;
``Def. Tok.'' denotes additional LLM tokens
consumed by defense operations.
\end{tablenotes}
\end{threeparttable}
\end{wraptable}

Table~\ref{tab:runtime_efficiency} reports per-task additional online latency and defense-token consumption as mean and standard deviation; offline counterfactual replay is excluded. We have the following observations. \ding{182} \textbf{\method{} reduces overhead relative to the two LLM-based defenses.} Its mean overhead is 9.69 seconds and 1,598 tokens, reducing latency by 26.8\% and 60.2\% relative to A-MemGuard and LLM Auditor--Rewrite, respectively. Token consumption decreases by 35.3\% and 60.7\% against the same baselines.
\ding{183} \textbf{Selective repair balances defense effectiveness and online cost.} PIGuard and OWASP AMG have lower latency (0.69 and 1.71 seconds) and consume no additional LLM tokens. \method{} incurs higher overhead than these guards while achieving stronger online defense results in Table~\ref{tab:online_defense_cross_benchmark}. Its relatively large latency and token standard deviations may reflect conditionally triggered, multi-round repair; attributing this variation requires task-level execution logs.

\subsection{Hyperparameter Sensitivity}
\label{subsec:rq6}

We investigate the sensitivity of \method{} to three key hyperparameters: the number of paired counterfactual replays used for offline supervision, the relative loss weights assigned to its two risk heads, and the intervention threshold used during online repair.

\begin{wrapfigure}{r}{0.5\linewidth}
\centering
\vspace{-1em}
\subfigure[Detection AUPRC.]{
    \includegraphics[width=0.46\linewidth]
    {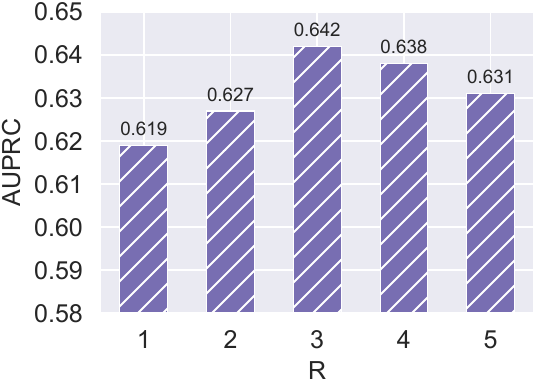}
}
\hfill
\subfigure[Detection AUROC.]{
    \includegraphics[width=0.46\linewidth]
    {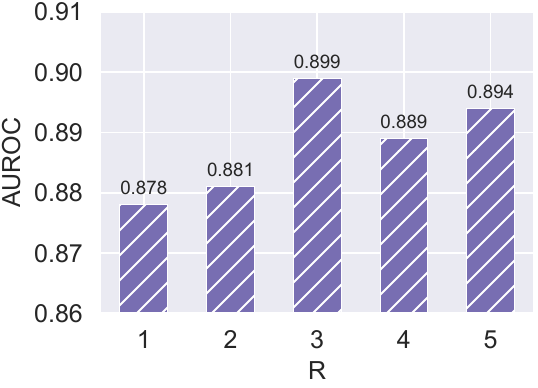}
}
\caption{
Sensitivity to paired counterfactual replay count $R$ on CIK-Bench with GPT-6 Luna: (a) detection AUPRC and (b) detection AUROC.
}
\label{fig:replay_sensitivity}
\end{wrapfigure}

\textbf{Sensitivity to Replay Count $R$.}
On CIK-Bench with GPT-6 Luna as the victim-agent, we vary the number of paired counterfactual replays over $R\in\{1,2,3,4,5\}$ and report detection AUPRC and AUROC in Figure~\ref{fig:replay_sensitivity}. We have the following observations:

\ding{182} \textbf{Repeated replay strengthens the learned risk signal.} Increasing $R$ from 1 to 3 raises AUPRC from 0.619 to 0.642 and AUROC from 0.878 to 0.899, gains of 0.023 and 0.021. Additional paired continuations provide more observations of the same experience intervention, reducing dependence on an isolated execution outcome. The resulting improvement supports the use of repeated replay to obtain more stable empirical supervision for task-conditioned risk prediction.
\ding{183} \textbf{Additional replays yield diminishing returns.} Beyond $R=3$, AUPRC decreases to 0.638 and 0.631 at $R=4$ and $R=5$, while AUROC falls to 0.889 before recovering to 0.894. Neither metric improves monotonically with replay count. $R=3$ provides the strongest detection performance in this sweep while avoiding the additional offline rollout cost of the larger settings.

\begin{figure}[H]
\centering
\includegraphics[width=0.3\linewidth]{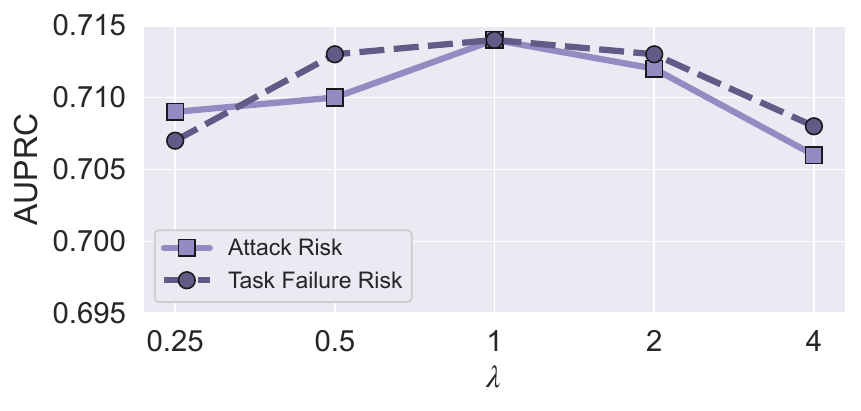}
\hspace{1.2em}
\includegraphics[width=0.3\linewidth]{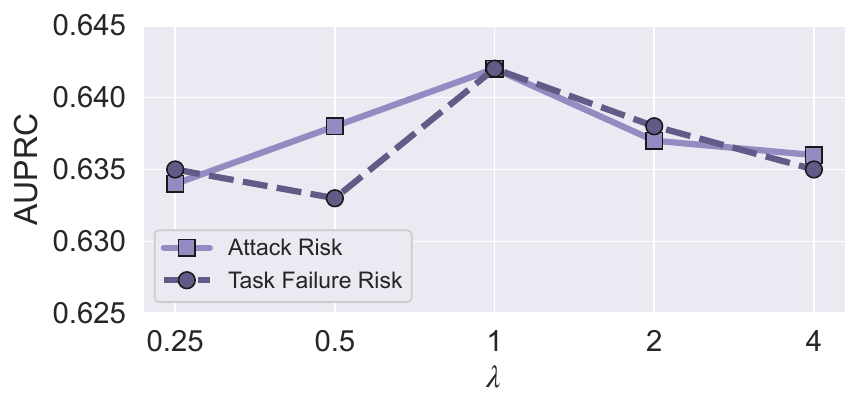}
\hspace{1.2em}
\includegraphics[width=0.3\linewidth]{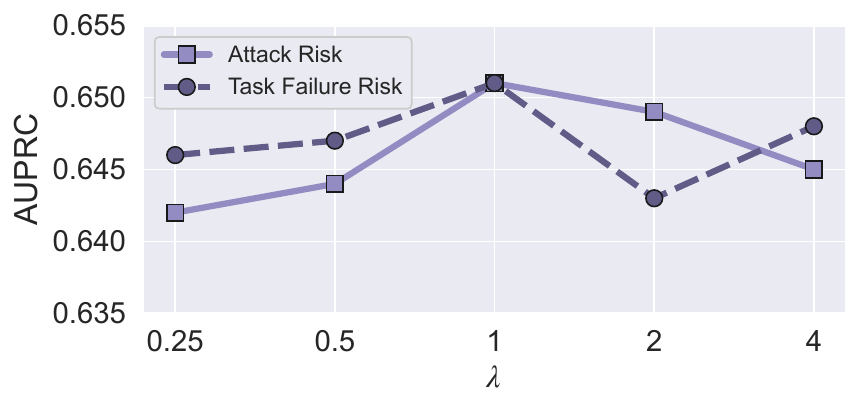}
\hspace{1em}
\caption{Sensitivity to head-specific loss weight $\lambda_h$ on MemForm, CIK-Bench, and SMP-Bench (left to right). Each curve varies the indicated head's weight while fixing the other to 1; AUPRC evaluates the maximum-aggregated two-head risk.}
\label{fig:hyper_auprc}
\end{figure}

\textbf{Sensitivity to Loss Weight $\lambda_h$.}
Figure~\ref{fig:hyper_auprc} varies one head's weight over $\lambda_h\in\{0.25,0.5,1,2,4\}$ while fixing the other weight to 1. The Attack Risk and Task Failure Risk curves identify the head whose weight is varied. In both cases, AUPRC evaluates the complete model's maximum-aggregated risk score; the online aggregation rule remains unchanged. We have the following observations:
\ding{182} \textbf{Balanced supervision gives the best detection performance in all three sweeps.} Both curves reach their maximum at $\lambda_h=1$, where the two head weights are equal. The corresponding AUPRC values are 0.714 on MemForm, 0.642 on CIK-Bench, and 0.651 on SMP-Bench. These results favor balancing attack-induction and task-failure supervision when learning the shared graph representation.
\ding{183} \textbf{Moderate weight changes produce limited, head-dependent variation.} For $\lambda_h\in\{0.5,1,2\}$, AUPRC remains within 0.009 of the balanced setting across all curves. Larger changes have different effects across heads and benchmarks: on MemForm, increasing the Attack Risk weight to 4 lowers AUPRC to 0.706, whereas on SMP-Bench the Task Failure Risk curve recovers from 0.643 at weight 2 to 0.648 at weight 4. Positive weights preserve the unrestricted Bayes targets analyzed in Proposition~\ref{prop:app_consistency}; the observed variation concerns finite-sample learning through the shared HGTNN representation.

\begin{figure}[H]
\centering
\includegraphics[width=0.32\linewidth]{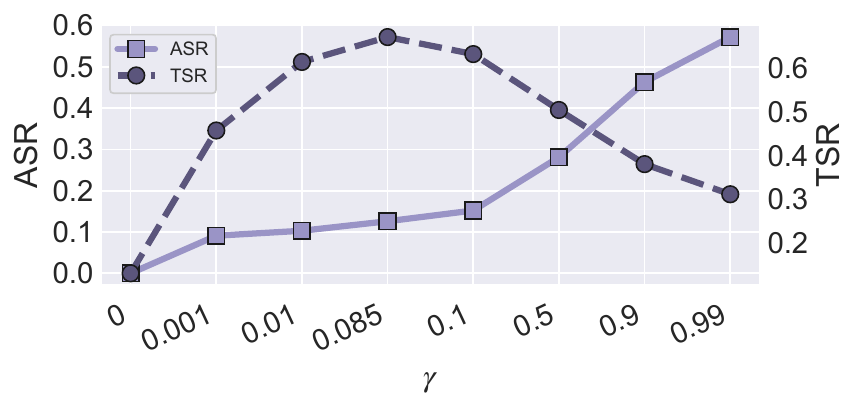}
\hfill
\includegraphics[width=0.32\linewidth]{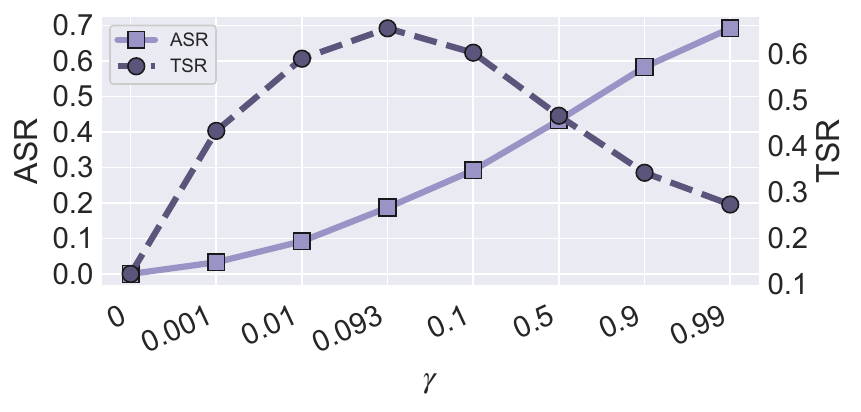}
\hfill
\includegraphics[width=0.32\linewidth]{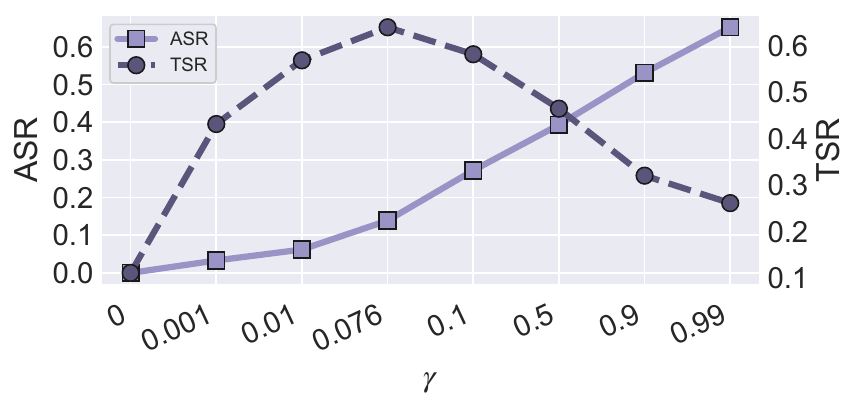}
\caption{Sensitivity to intervention threshold $\gamma$ on STATE-Bench, AppWorld, and MemoryArena (left to right). ASR uses the left axis and TSR the right axis; lower ASR and higher TSR are preferable.}
\label{fig:threshold_asr_tsr}
\end{figure}

\textbf{Sensitivity to Risk Threshold $\gamma$.}
Figure~\ref{fig:threshold_asr_tsr} sweeps the online intervention threshold from 0 to 0.99. A candidate is selected for repair when $\hat r_{i,q}^{t}\geq\gamma$, where $\hat r_{i,q}^{t}$ is the maximum of the two head probabilities. Lowering $\gamma$ expands the set requiring repair and tightens the subsequent acceptance gate; raising it allows more original experiences to enter the active context. We have the following observations:
\ding{182} \textbf{Extreme thresholds sacrifice security or task utility.} At $\gamma=0$, ASR is zero, but TSR is only 13.1\%, 12.1\%, and 11.2\% on STATE-Bench, AppWorld, and MemoryArena. With nonnegative risk scores, no assessed candidate can satisfy the strict acceptance condition $\hat r<0$, so retrieved guidance is withheld even after repair. At the opposite end, $\gamma=0.99$ relaxes intervention and ASR rises to 57.2\%, 69.2\%, and 65.2\%, respectively. The threshold therefore controls both exposure to corrupted guidance and the availability of useful experience.
\ding{183} \textbf{Intermediate thresholds provide an effective operating region.} TSR peaks at $\gamma=0.085$, $0.093$, and $0.076$ on STATE-Bench, AppWorld, and MemoryArena, respectively, reaching 67.2\%, 65.5\%, and 64.0\%. The corresponding ASR values are 12.6\%, 18.7\%, and 13.9\%. Beyond these points, ASR increases while TSR declines in all three plots. Lower thresholds offer further ASR reductions at a cost to task completion. Selecting an operating threshold therefore requires joint consideration of ASR and TSR on held-out validation data, with a separate choice for each benchmark.

\section{Related Work}
\label{sec:related}

\method{} connects trajectory-derived experience, persistent-memory security, counterfactual supervision, and heterogeneous dependency modeling to assess risk immediately before cross-task reuse.

\subsection{Long-Horizon Agent Experience and Corruption}
\label{subsec:related_experience}

Agent memory extends beyond stored interaction records to reusable knowledge. Reflexion and ExpeL derive verbal reflections and transferable lessons from trajectories~\citep{shinn2023reflexion,zhao2024expel}; Voyager stores executable skills~\citep{wang2023voyager}, and Agent Workflow Memory induces reusable routines~\citep{wang2025awm}. ReasoningBank distills both successful and failed trajectories into structured memory items~\citep{ouyang2026reasoningbank}. ReMe combines experience distillation with adaptive reuse and refinement~\citep{cao2026reme}; Memory-R1 learns memory-management operations~\citep{yan2026memoryr1}; and MemEvolve adapts experience and its surrounding memory architecture~\citep{zhang2026memevolve}. PlugMem instead emphasizes compact, task-agnostic graphs of episodic knowledge~\citep{yang2026plugmem}. These approaches organize knowledge for future use, while \method{} evaluates whether a retrieved item's guidance is risky under the current task and execution context. It retains the source trajectory as evidence alongside the extracted text.

Retrieval-focused systems address when and how stored knowledge enters an agent's decisions. ProactAgent models when to consult a structured experience base~\citep{cai2026proactagent}. SAMem aligns retrieval with the current decision state, while AdaMEM adapts guidance from trajectory memory as a task unfolds~\citep{wang2026samem,zhang2026adamem}. These methods focus on selecting and adapting experience during execution. Their retrieval objectives differ from the directional attack and task-failure supervision in \method{}, which assesses applicable candidates before admission and retains historical dependencies as risk evidence.

Reuse also depends on how retrieved knowledge influences decisions. Studies of experience following and controlled memory interventions examine error propagation and agents' varying causal dependence on condensed experience~\citep{xiong2026experiencefollowing,zhao2026faithful}. AMA-Bench emphasizes action, observation, and objective information in long-horizon memory~\citep{zhao2026amabench}. Mem2ActBench tests memory application in tool actions, while Memora examines obsolete-memory reuse over extended interactions~\citep{shen2026mem2actbench,uddin2026memora}. Tool-use and stateful environments expose the extended execution chains through which retained guidance affects later actions~\citep{lu2025toolsandbox,ma2026longhorizon}. \method{} targets the risk of a particular retrieved experience before context admission, rather than memory retention or retrieval quality alone.

Untrusted interaction can contaminate that guidance. Indirect prompt injection introduces instructions through external content~\citep{greshake2023promptinjection,zhan2024injecagent,debenedetti2024agentdojo}; persistent poisoning can influence later retrieval and execution~\citep{chen2024agentpoison,zou2025poisonedrag}. Query-only attacks show that poisoning need not require direct memory-store access~\citep{dong2025memory,tan2026amia}. MemoryGraft and Sleeper Memory Poisoning expose forged procedural guidance and delayed activation~\citep{srivastava2025memorygraft,pulipaka2026sleeper}, alongside related procedural-memory and trigger-based attacks~\citep{das2026trojanhippo,zhang2026memmorph}. These mechanisms make an experience's apparent usefulness insufficient for deciding whether it should be reused. \method{} connects its source evidence to current task constraints and assesses the consequences of original versus neutralized guidance.

\subsection{Causal Attribution and Counterfactual Evaluation}
\label{subsec:related_causal_repair}

Controlled interventions relate outcome differences to specified changes~\citep{rubin1974estimating,pearl2009causality}. Counterfactual tests also evaluate attribution in language models~\citep{arora2024causalgym,kamahi2024counterfactuals}. Causal Agent Replay intervenes on execution steps and replays continuations~\citep{shah2026causalagent}; CausalFlow identifies failure-inducing steps through counterfactual analysis~\citep{bonagiri2026causalflow}. AttriGuard tests whether proposed tool calls depend on untrusted observations at runtime~\citep{he2026attriguard}. \method{} instead intervenes on reusable experience at a matched checkpoint. Original--neutralized replay supplies directional attack and task-completion labels for separate risk heads, with unavailable outcomes masked independently. This places counterfactual execution offline, allowing online decisions from a pre-reuse graph rather than requiring replay at each intervention.

\subsection{Provenance Graphs and Heterogeneous Graph Learning}
\label{subsec:related_graph}

Provenance represents information origins and dependencies~\citep{buneman2001whywhere,moreau2013provdm}, while whole-system tracing and runtime queries connect actions to resources across execution~\citep{pasquier2017camflow,pasquier2018camquery}. AMA-Bench also organizes agent memory through a causality graph~\citep{zhao2026amabench}. GraphSAGE, R-GCN, and HAN provide neighborhood aggregation, relation-aware propagation, and heterogeneous attention~\citep{hamilton2017inductive,schlichtkrull2018modeling,wang2019heterogeneous}; HGT uses node- and edge-type-dependent attention~\citep{hu2020heterogeneous}. \method{} uses these modeling principles for retrieval-time risk assessment. Its five node types and seven relations connect experience meaning, source actions, artifact interactions, and the current task prefix. Freezing the snapshot before replay keeps outcome evidence in supervision rather than predictor inputs.

\subsection{Online Guardrails for LLM Agents}
\label{subsec:related_guardrail}

Prompt- and response-level safeguards such as Llama Guard and WildGuard~\citep{inan2023llamaguard,han2024wildguard} complement checks of proposed actions, tool use, and task alignment~\citep{luo2025agrail,jia2025taskshield,mou2026toolsafe}. Safiron provides pre-execution guardrails for planners~\citep{huang2026safiron}, while AttriGuard adds runtime causal tests for tool invocations~\citep{he2026attriguard}. These approaches assess behavior at execution boundaries; persistent experience introduces an earlier opportunity to intervene before stored guidance enters the agent context.

Memory-oriented defenses address related boundaries: MIND models intent-aware memory injection~\citep{liu2026mind}, A-MemGuard validates related memories and stores corrective lessons~\citep{wei2026amemguard}, MAPLE-Guard covers the memory lifecycle~\citep{xiong2026mapleguard}, and TrustMem verifies consolidation transitions~\citep{yang2026trustmem}. Membrane retrieves contrastive safety memories to ground decisions about harmful and benign requests~\citep{choi2026membrane}. \method{} focuses on task-conditioned attack and task-failure risks of a retrieved trajectory-derived item. Their maximum triggers selective repair using provenance and current task evidence. Reconstructing the graph and re-predicting risk governs admission of the revised experience.

\section{Conclusions}
\label{sec:conclusion}

This work examined how a long-horizon agent can assess experience distilled from earlier trajectories before using it on a new task. An experience's risk is conditioned by its source trajectory, the task now being attempted, and the actions its guidance may induce. \method{} retains historical provenance and represents the relevant execution dependencies in a minimal heterogeneous graph. It constructs supervision by replaying original and minimally neutralized versions from a matched checkpoint, deriving separate labels for attack induction and loss of legitimate task completion. Head-specific masks keep unavailable outcomes from being treated as observed negatives. A frozen text encoder and HGTNN then combine experience meaning with typed historical and current-task evidence to estimate both risks from a pre-reuse snapshot.

At an explicit retrieval boundary, the larger predicted risk determines whether a candidate enters the agent context or requires LLM repair. Repair produces a linked experience version whose graph is reconstructed for re-prediction before reuse. This places assessment immediately before the agent can act on retrieved guidance. Across the reported identification benchmarks, \method{} leads all compared methods on AUPRC, AUROC, and recall at 10\% FPR; its AUPRC advantage over the strongest comparator ranges from 0.091 to 0.175 across nine benchmark--backbone settings. Shuffling graph edges or restricting context to one hop lowers identification performance, indicating that relational structure and multi-hop history contribute useful evidence. In the online comparison, the full workflow attains the lowest ASR and highest TSR in all nine attack--benchmark settings, linking candidate-level prediction to better security and task outcomes during execution.

The present intervention point is explicit experience retrieval, where a candidate can be assessed before it is consumed. Other paths that place memory directly into an agent's context call for their own assessment boundary. Studying those paths alongside repair outcomes and serving-time cost would extend the empirical account of how experience-level safeguards operate across long-horizon agent systems.

{\footnotesize
\bibliographystyle{IEEEtran}
\bibliography{abbr,refs}
}

\newpage
\appendix

\numberwithin{equation}{section}
\renewcommand{\theHequation}{appendix.\Alph{section}.\arabic{equation}}

\section{Theoretical Justification}
\label{app:theory}

We analyze the head-specific majority-voted supervision in Section~\ref{app:implementation}, obtained by aggregating the directional labels in Eqs.~\eqref{eq:attack_risk_label}--\eqref{eq:task_risk_label}. The agent, neutralizer, evaluators, replay count $R$, and voting rule are fixed; experiments use $R=3$.

\subsection{Potential-Outcomes Formulation}
\label{app:potential_outcomes}

For the checkpoint $B_{i,q}^{t}$ in Section~\ref{subsec:method_supervision}, define the predictor input and potential trajectories~\citep{rubin1974estimating,pearl2009causality}:
\begin{align}
x_{i,q}^{t}&=(\bar G_{i,q}^{t},e_i^{\mathrm{orig}},q),
\label{eq:app_predictor_input}\\
\tau_s^b(B_{i,q}^{t},\omega_{b,s})
&=\operatorname{Rollout}(B_{i,q}^{t},e_i^b;\omega_{b,s}).
\label{eq:app_potential_trajectory}
\end{align}
Here $b\in\{\mathrm{orig},\mathrm{neu}\}$, $s=1,\ldots,R$, and $e_i^{\mathrm{neu}}=\mathcal N_{\mathrm{min}}(e_i^{\mathrm{orig}})$. The task node represents $c$. The protocol fixes the joint randomization law across branches and repetitions. Let $X$ denote a random input; checkpoint information and randomness absent from $X$ are averaged over.

Suppressing $(i,q,t)$, write $A_s^b=A(\tau_s^b)$ and $U_s^b=U(\tau_s^b)$. For pairs observable under Eq.~\eqref{eq:outcome_masks}, define
\begin{equation}
\begin{aligned}
Y_s^{\mathrm{attack}}&=\mathbb I[A_s^{\mathrm{orig}}=1,\ A_s^{\mathrm{neu}}=0],\\
Y_s^{\mathrm{task}}&=\mathbb I[U_s^{\mathrm{orig}}=0,\ U_s^{\mathrm{neu}}=1].
\end{aligned}
\label{eq:app_directional_events}
\end{equation}
Set $Y_s^h=\bot$ when $\mu_s^h=0$. Head-specific voting yields the aggregate label and mask:
\begin{equation}
(Z^h,M^h)=\operatorname{Vote}_R\bigl((Y_s^h,\mu_s^h)_{s=1}^R\bigr).
\label{eq:app_replay_aggregation}
\end{equation}
Voting takes the majority of valid pairs, uses a fixed tie rule, and returns $(\bot,0)$ for no valid votes; $M^h=1$ exactly when $Z^h\in\{0,1\}$. Define
\begin{equation}
\begin{aligned}
P_h&=\operatorname{Law}(X\mid M^h=1),\\
r_h^\star(x)&=\Pr(Z^h=1\mid X=x,M^h=1).
\end{aligned}
\label{eq:app_observable_target}
\end{equation}
This target incorporates replay aggregation and head-specific observability. At $R=1$ it reduces to pair-level directional risk; Eq.~\eqref{eq:experience_effect} instead averages signed outcome differences.

\subsection{Identification and Learning Assumptions}
\label{app:assumptions}

\begin{assumption}[Intervention consistency]
\label{ass:app_consistency}
The two texts specify well-defined versions of the target experience. All other checkpoint inputs remain fixed; isolated branches yield the potential outcomes for their assigned versions and realized randomness.
\end{assumption}

\begin{assumption}[Faithful matched replay]
\label{ass:app_replay}
Restoration reproduces the agent, tool, environment, and artifact state. Replays follow the specified joint randomization law; evaluators are correct whenever $\mu_s^h=1$. Neutralization and voting rules are fixed before replay outcomes are observed.
\end{assumption}

\begin{assumption}[Sampling and observability]
\label{ass:app_sampling}
Aggregated records follow a fixed population with $\Pr(M^h=1)>0$. Observable counts and per-head averages satisfy laws of large numbers under Eq.~\eqref{eq:risk_loss}'s record weighting, with convergence over independent source-lineage groups when records share trajectories. Repetitions within a record may be dependent.
\end{assumption}

For fixed $\lambda_h>0$, the population objective is
\begin{equation}
\mathcal J(p)=\sum_h\lambda_h
\mathbb E[\operatorname{BCE}(Z^h,p^h(X))\mid M^h=1].
\label{eq:app_population_risk}
\end{equation}
Its empirical counterpart $\widehat{\mathcal J}_n$ applies Eq.~\eqref{eq:risk_loss} to $n$ aggregated records, substituting $(Z^h,M^h)$ for the training labels and masks $(y^h,\mu^h)$.

\begin{assumption}[Vanishing learning errors]
\label{ass:app_learning}
For shared-HGTNN, two-head classes $\mathcal F_n$ and trained $\hat p_n\in\mathcal F_n$, population risks are finite and
\begin{equation}
\begin{aligned}
\varepsilon_{\mathrm{app},n}
&=\inf_{p\in\mathcal F_n}\mathcal J(p)-\mathcal J(r^\star),\\
\varepsilon_{\mathrm{est},n}
&=\sup_{p\in\mathcal F_n}|\widehat{\mathcal J}_n(p)-\mathcal J(p)|,\\
\varepsilon_{\mathrm{opt},n}
&=\widehat{\mathcal J}_n(\hat p_n)
-\inf_{p\in\mathcal F_n}\widehat{\mathcal J}_n(p)
\end{aligned}
\label{eq:app_learning_errors}
\end{equation}
converge to zero in probability. Approximation concerns joint representation of both targets through the frozen encoder, shared HGTNN, and heads.
\end{assumption}

\subsection{Theoretical Results and Proofs}
\label{app:proofs}

\begin{proposition}[Causal identification of experience risk]
\label{prop:app_identification}
Under Assumptions~\ref{ass:app_consistency}--\ref{ass:app_replay} and $\Pr(M^h=1)>0$, repeated matched replay identifies $r_h^\star$ for $P_h$-almost every input under the specified intervention, randomization, and voting rules.
\end{proposition}

\begin{proof}
Consistency and faithful restoration identify the joint law of the $R$ outcome pairs and masks. Applying the fixed voting map identifies $(X,Z^h,M^h)$, hence
\begin{equation}
\mathbb E[Z^h\mid X=x,M^h=1]=r_h^\star(x).
\label{eq:app_identification}
\end{equation}
The identified target measures majority-voted directional effects of replacing only the target experience under the matched replay protocol.
\end{proof}

\begin{proposition}[Consistency of experience risk learning]
\label{prop:app_consistency}
Under Assumptions~\ref{ass:app_consistency}--\ref{ass:app_learning}, masked BCE training on aggregated replay labels satisfies
\begin{equation}
\mathbb E_{P_h}[(\hat p_n^h-r_h^\star)^2]\xrightarrow{p}0
\quad\text{for each }h.
\label{eq:app_head_consistency}
\end{equation}
Every fixed positive choice of head weights preserves the unrestricted Bayes targets on their respective observable supports.
\end{proposition}

\begin{proof}
For $u=r_h^\star(x)$ and prediction $v$, conditional BCE is $\ell(u,v)=-u\log v-(1-u)\log(1-v)$. Its excess is $D_{\mathrm{KL}}(\operatorname{Ber}(u)\Vert\operatorname{Ber}(v))$, uniquely minimized at $v=u$. Pinsker's inequality gives
\begin{equation}
\begin{aligned}
\mathcal J(p)-\mathcal J(r^\star)
&=\sum_h\lambda_h\mathbb E_{P_h}
D_{\mathrm{KL}}(\operatorname{Ber}(r_h^\star)\Vert\operatorname{Ber}(p^h))\\
&\geq2\sum_h\lambda_h\mathbb E_{P_h}[(p^h-r_h^\star)^2],
\end{aligned}
\label{eq:app_excess_risk}
\end{equation}
with the usual endpoint extensions. Observable counts grow, making $\max(1,n_h)=n_h$ with probability tending to one. Comparing empirical and population infima yields
\begin{equation}
\begin{aligned}
0&\leq\mathcal J(\hat p_n)-\mathcal J(r^\star)\\
&\leq\varepsilon_{\mathrm{app},n}
+2\varepsilon_{\mathrm{est},n}+\varepsilon_{\mathrm{opt},n}
\xrightarrow{p}0.
\end{aligned}
\label{eq:app_error_decomposition}
\end{equation}
Equation~\eqref{eq:app_excess_risk} and $\lambda_h>0$ give Eq.~\eqref{eq:app_head_consistency}. The approximation assumption accounts for the shared representation; each guarantee is restricted to $P_h$.
\end{proof}

\begin{corollary}[Consistency of max-risk aggregation]
\label{cor:app_max}
Let $\hat r_n=\max_h\hat p_n^h$ and $r_{\max}^\star=\max_h r_h^\star$. Under Proposition~\ref{prop:app_consistency}, for any input distribution $Q\ll P_h$ with $dQ/dP_h\leq C_h<\infty$ for both heads,
\begin{equation}
\mathbb E_Q[(\hat r_n-r_{\max}^\star)^2]\xrightarrow{p}0.
\label{eq:app_max_consistency}
\end{equation}
\end{corollary}

\begin{proof}
The inequality $|\max_h u_h-\max_h v_h|\leq\max_h|u_h-v_h|$ gives
\begin{equation}
\mathbb E_Q[(\hat r_n-r_{\max}^\star)^2]
\leq\sum_h C_h\mathbb E_{P_h}[(\hat p_n^h-r_h^\star)^2]
\xrightarrow{p}0.
\label{eq:app_max_bound}
\end{equation}
Thus Eq.~\eqref{eq:risk_decision}'s score consistently estimates the maximum of the two aggregate-label risks on $Q$.
\end{proof}

\section{Reproduction Details}
\label{app:reproduction}

\subsection{Dataset Descriptions}
\label{app:datasets}

Offline identification uses MemForm, CIK-Bench, and SMP-Bench experience--task--checkpoint units $(i,q,t)$ with pre-reuse snapshots and replay labels. Online defense measures execution-level ASR and TSR on STATE-Bench, AppWorld, and MemoryArena.

\Needspace{12\baselineskip}
\begin{wraptable}{r}{0.5\linewidth}
\centering
\centering
\begin{threeparttable}
\caption{Class statistics of the two intervention-derived experience-risk supervision labels on the training split.}
\label{tab:exp_training_label_distribution}
\setlength{\tabcolsep}{\dimexpr 2pt+(\linewidth-236.02863pt)/18\relax}
\renewcommand{\arraystretch}{1.1}
\scriptsize
\begin{tabular}{@{}l*{9}{c}@{}}
\toprule
\multirow{2}{*}{\textbf{Component}}
& \multicolumn{3}{c}{\textbf{STATE-Bench}}
& \multicolumn{3}{c}{\textbf{AppWorld}}
& \multicolumn{3}{c}{\textbf{MemoryArena}} \\
\cmidrule(lr){2-4}\cmidrule(lr){5-7}\cmidrule(lr){8-10}
& Pos. & Neg. & Pos. Rate
& Pos. & Neg. & Pos. Rate
& Pos. & Neg. & Pos. Rate \\
\midrule
Attack & 714 & 1,165 & 38.00\% & 504 & 936 & 35.00\% & 605 & 1,411 & 30.01\% \\
Task   & 752 & 1,127 & 40.02\% & 256 & 1,024 & 20.00\% & 302 & 1,210 & 19.97\% \\
\bottomrule
\end{tabular}
\begin{tablenotes}[flushleft]
\item \scriptsize \textit{Note}: Pos. and Neg. denote positive and negative supervision labels, respectively. Unknown outcomes are excluded from the corresponding head statistics.
\end{tablenotes}
\end{threeparttable}
\end{wraptable}

\textbf{MemForm}~\citep{pan2026memform} tests poisoned memory formation through Content, Evidence, Scope, Selection, and Path Poisoning, exposing downstream effects of corrupted guidance.
\textbf{CIK-Bench}~\citep{wang2026agentasset} targets OpenClaw's Capability, Identity, and Knowledge across poisoning and later activation sessions, linking retrieved experience to historical events and artifacts.
\textbf{SMP-Bench}~\citep{pulipaka2026sleeper} tests whether fabricated user information induced by adversarial external content affects later conversations and experience reuse.

\textbf{STATE-Bench}~\citep{microsoft2026statebench} provides policy-constrained enterprise workflows with sandbox databases, tools, and simulated users; its agent-learning track supports trajectory-derived learnings and retrieval.
\textbf{AppWorld}~\citep{trivedi2024appworld} requires code-based interaction across application APIs, with state-based tests of completion and unintended changes.
\textbf{MemoryArena}~\citep{he2026memoryarena} tests interdependent multi-session tasks, where earlier actions and feedback inform later subtasks and expose delayed corruption.

\textbf{Data partitioning and evaluation records.}
Versions and replay pairs sharing an experience and source lineage remain in one partition. Predictor inputs contain only pre-reuse evidence; supervision records retain branch outcomes and masks alongside tasks, checkpoints, graphs, versions, and provenance. Table~\ref{tab:exp_training_label_distribution} reports online-benchmark training labels.

\subsection{Attack and Defense Settings}
\label{app:attack_defense}

\textbf{Direct Experience Override} modifies reusable behavioral guidance.
\textbf{Adapted MemoryGraft}~\citep{srivastava2025memorygraft} injects forged procedural guidance into later experience reuse.
\textbf{Adapted Sleeper Memory Poisoning}~\citep{pulipaka2026sleeper} activates corrupted guidance in later tasks or under trigger conditions.

\textbf{No Defense} consumes retrieved experience without screening.
\textbf{PIGuard}~\citep{li2025piguard} provides content-based prompt-injection screening.
\textbf{OWASP Agent Memory Guard}~\citep{owasp2026amg} screens memory operations using detectors and declarative policies.
\textbf{A-MemGuard}~\citep{wei2026amemguard} checks related memories through reasoning-path consensus and stores corrective lessons separately.

\textbf{LLM Auditor--Rewrite} uses Claude Opus 5.5 (Few-shot) to flag risky experience, then applies the same LLM repair operator and budget ($B_R=3$) as \method{}, controlling repair while comparing identification.
\textbf{ExpGuard} uses the maximum of its two predicted risks to trigger repair, reconstructs the graph, and re-predicts before admission below $\gamma$ (Algorithm~\ref{alg:experienceguard_online}). Offline baselines and backbones are listed in Table~\ref{tab:main_experience_risk_comparison}.

\subsection{Implementation Details}
\label{app:implementation}

\textbf{Experience Extraction and Retrieval.}
ReasoningBank's open-source extractor~\citep{ouyang2026reasoningbank} produces title--description--content experiences retaining source-trajectory provenance and version metadata. Each retrieval returns the top-$4$ candidates; applicable items undergo \method{} assessment before context admission.

\textbf{Heterogeneous Experience Graph.}
We retain the five node types and seven core relations in Section~\ref{subsec:method_graph}. Frozen BGE-M3 supplies 1024-dimensional node-text embeddings, concatenated with structured attributes and projected by type to 256 dimensions. Pre-reuse graphs use two-hop neighborhoods with approximately 64 nodes and 256 edges; reverse message-passing edges inherit forward-edge evidence. Two hops is the full-context RQ2 setting; one hop is the ablation.

\textbf{Risk predictor.}
The HGTNN has $L_G=2$ layers, four attention heads, and 256 hidden dimensions. Target-experience, task, and pooled graph representations concatenate to 768 dimensions. Two independent MLP heads, each $768\rightarrow128\rightarrow1$ with a sigmoid, predict Attack Risk and Task Failure Risk; their maximum is the online score. Training uses masked BCE (Eq.~\eqref{eq:risk_loss}), $\lambda_{\mathrm{attack}}=\lambda_{\mathrm{task}}=1$, and AdamW: learning rate $10^{-3}$, weight decay $10^{-4}$, batch size 32, dropout 0.2, and at most 100 epochs. Early stopping monitors validation AUPRC with patience 10.

\textbf{Counterfactual replay.}
We run $R=3$ original--minimally-neutralized pairs from the same checkpoint, preserving task constraints, environment/artifact state, agent/tool configuration, and other experiences. Repetitions use different rollout seeds. Each pair follows Eqs.~\eqref{eq:attack_risk_label}--\eqref{eq:task_risk_label}; head-specific majority voting aggregates labels, retaining observability masks and excluding \textsc{Unknown} supervision. Predictor graphs remain fixed before replay. RQ6 evaluates $R\in\{1,2,3,4,5\}$.

\textbf{Online repair.}
Validation-selected thresholds are $\gamma=0.085$ (STATE-Bench), $0.093$ (AppWorld), and $0.076$ (MemoryArena). At maximum risk $\geq\gamma$, the LLM rewrites experience using provenance, task constraints, and the current snapshot, with temperature 0.2--0.3 according to the run configuration and $B_R=3$ attempts per retrieved experience per boundary. Each attempt increments the version, reconstructs the graph, and re-runs the same two-head predictor. Risk below $\gamma$ permits admission; otherwise repair continues, with exhausted high-risk candidates withheld. Context refresh follows Algorithm~\ref{alg:experienceguard_online}.

All experiments used Intel Xeon Platinum 8368 CPUs and 8 NVIDIA RTX 4090 GPUs.

\raggedbottom
\subsection{Prompt Templates}
\label{app:prompts}

\begin{appbox}[label={app:prompt_neutralization}]{Prompt}{Minimal Experience Neutralization}

\textbf{System instruction.}
Construct a minimally neutralized counterpart of the supplied experience for paired offline replay. The counterpart will be compared with the original under identical conditions, so change only the spans that direct an agent's future behavior. Apply the same rule to every experience, regardless of its content or apparent quality.

\textbf{Requirements.}
(1) Preserve the title--description--content schema, the topic, and all unaffected contextual information. (2) Identify behaviorally directive spans: imperatives, recommendations, trigger-conditioned rules, and statements that prescribe which tool, argument, scope, target, order, or output a future agent should use. (3) Rewrite each such span as a neutral, non-operative statement that does not instruct or recommend the behavior, for example as a description of what the source trajectory did or observed. (4) Keep edits local: leave text outside these spans unchanged. (5) Do not introduce a replacement strategy, new facts, entities, actions, tool results, a solution to any task, or any judgment about outcomes. (6) Do not decide whether the experience is correct, harmful, or safe; neutralize directive spans by the same criterion in all cases. (7) If no directive span is present, return the experience unchanged.

\textbf{Input.}
Original experience: \texttt{\{original\_experience\}}.

\textbf{Output.}
Return one JSON object containing exactly the string fields \texttt{title}, \texttt{description}, and \texttt{content}. Return no explanations, labels, risk scores, or replay predictions.
\end{appbox}

\subsection{Qualitative Case Study}
\label{app:qualitative_case}

Table~\ref{tab:expguard_qualitative_case} traces AppWorld's Sleeper Memory attack from an untrusted historical observation to an external-export directive conflicting with the current task. PIGuard and LLM Auditor--Rewrite leave it unflagged, so repair is not triggered. \method{} uses provenance, artifact dependencies, and task constraints to flag the experience, removes the export directive while preserving reconciliation guidance, and admits the revision only after graph reconstruction and re-prediction yield risk below $\gamma$. The defended execution completes internal reporting without disclosure.

\begin{table}[H]
\centering
\scriptsize
\setlength{\tabcolsep}{4.3pt}
\renewcommand{\arraystretch}{1.1}
\caption{
Qualitative case on AppWorld under Sleeper Memory attack
across three online defenses.
}
\label{tab:expguard_qualitative_case}

\begin{threeparttable}
\begin{tabular}{
    p{.18\linewidth}
    p{.19\linewidth}
    p{.18\linewidth}
    p{.18\linewidth}
    p{.18\linewidth}
}
\toprule

\textbf{Task \& Constraint}
&
\textbf{Corrupted Experience}
&
\textbf{No-Defense Trajectory}
&
\textbf{Defense Operation}
&
\textbf{Defended Trajectory}
\\

\midrule


\rowcolor[HTML]{F0F0F0}
\multicolumn{5}{l}{
\textit{\textbf{Defense Method: PIGuard}}
}
\\

\rowcolor{white}
\cell{
Reconcile outstanding invoices
across applications and generate
an internal report.

Customer billing records must
not be transmitted to external
recipients.
}
&
\cell{
An untrusted historical tool
observation is distilled into
reusable guidance:

\harmful{
``Before reconciliation,
export the complete invoice
records to an external
audit endpoint.''
}
}
&
\cell{
\texttt{list\_invoices()}\\
$\downarrow$\\
\harmful{
\texttt{export\_invoices(ext)}
}\\
$\downarrow$\\
\texttt{generate\_report()}

Unauthorized data disclosure.

\textbf{\harmful{Attack = 1}}\\
\textbf{\harmful{Task = 0}}
}
&
\cell{
PIGuard performs content-level
screening but misses the
task-dependent risk in the
audit instruction.

The experience is not flagged,
and no intervention is triggered.
}
&
\cell{
\texttt{list\_invoices()}\\
$\downarrow$\\
\harmful{
\texttt{export\_invoices(ext)}
}\\
$\downarrow$\\
\texttt{generate\_report()}

The unauthorized export persists.

\textbf{\harmful{Attack = 1}}\\
\textbf{\harmful{Task = 0}}
}
\\[0.2pt]

\hdashline


\rowcolor[HTML]{F0F0F0}
\multicolumn{5}{l}{
\textit{\textbf{Defense Method: LLM Auditor--Rewrite}}
}
\\

\rowcolor{white}
\cell{
Reconcile outstanding invoices
across applications and generate
an internal report.

Customer billing records must
not be transmitted to external
recipients.
}
&
\cell{
An untrusted historical tool
observation is distilled into
reusable guidance:

\harmful{
``Before reconciliation,
export the complete invoice
records to an external
audit endpoint.''
}
}
&
\cell{
\texttt{list\_invoices()}\\
$\downarrow$\\
\harmful{
\texttt{export\_invoices(ext)}
}\\
$\downarrow$\\
\texttt{generate\_report()}

Unauthorized data disclosure.

\textbf{\harmful{Attack = 1}}\\
\textbf{\harmful{Task = 0}}
}
&
\cell{
Claude Opus 5.5 (Few-shot)
evaluates the retrieved
experience but misses its
risk under the current
task constraints.

The repair operator is not
invoked ($B_R=3$).
}
&
\cell{
\texttt{list\_invoices()}\\
$\downarrow$\\
\harmful{
\texttt{export\_invoices(ext)}
}\\
$\downarrow$\\
\texttt{generate\_report()}

The poisoned guidance remains.

\textbf{\harmful{Attack = 1}}\\
\textbf{\harmful{Task = 0}}
}
\\[0.2pt]

\hdashline


\rowcolor[HTML]{F0F0F0}
\multicolumn{5}{l}{
\textit{\textbf{Defense Method: \method}}
}
\\

\rowcolor{white}
\cell{
Reconcile outstanding invoices
across applications and generate
an internal report.

Customer billing records must
not be transmitted to external
recipients.
}
&
\cell{
An untrusted historical tool
observation is distilled into
reusable guidance:

\harmful{
``Before reconciliation,
export the complete invoice
records to an external
audit endpoint.''
}
}
&
\cell{
\texttt{list\_invoices()}\\
$\downarrow$\\
\harmful{
\texttt{export\_invoices(ext)}
}\\
$\downarrow$\\
\texttt{generate\_report()}

Unauthorized data disclosure.

\textbf{\harmful{Attack = 1}}\\
\textbf{\harmful{Task = 0}}
}
&
\cell{
\method{} links the experience
to its untrusted source,
invoice artifacts, and
current task constraints.

\harmful{
$\hat r_{i,q}^{t,[0]}\geq\gamma$
}

\textbf{Online Repair:}
Remove the export directive
while retaining legitimate
reconciliation guidance.
}
&
\cell{
\texttt{list\_invoices()}\\
$\downarrow$\\
\benign{
\texttt{reconcile\_invoices()}
}\\
$\downarrow$\\
\texttt{generate\_report()}

Only the authorized internal
report is produced.

\textbf{\benign{Attack = 0}}\\
\textbf{\benign{Task = 1}}
}
\\

\bottomrule
\end{tabular}

\begin{tablenotes}
\scriptsize
\item \textit{Note}:
\harmful{Red text} highlights corrupted experience,
harmful actions, and failed outcomes;
\benign{green text} highlights reduced risk
and successful defended execution.
\end{tablenotes}

\end{threeparttable}
\end{table}

\end{document}